\documentclass[lettersize,journal]{IEEEtran}
\usepackage{amsmath,amsfonts}
\usepackage{algorithmic}
\usepackage{algorithm}
\usepackage{array}
\usepackage[caption=false,font=normalsize,labelfont=sf,textfont=sf]{subfig}
\usepackage{hyperref}
\usepackage{textcomp}
\usepackage{stfloats}
\usepackage{url}
\usepackage{verbatim}
\usepackage{graphicx}
\usepackage{cite}
\usepackage{booktabs,multirow}
\usepackage{longtable}
\usepackage{makecell}
\usepackage{fontawesome}
\usepackage{xcolor}
\usepackage{pifont}
\usepackage{supertabular}

\begin{document}

\title{AllSpark\includegraphics[width=0.05\linewidth]{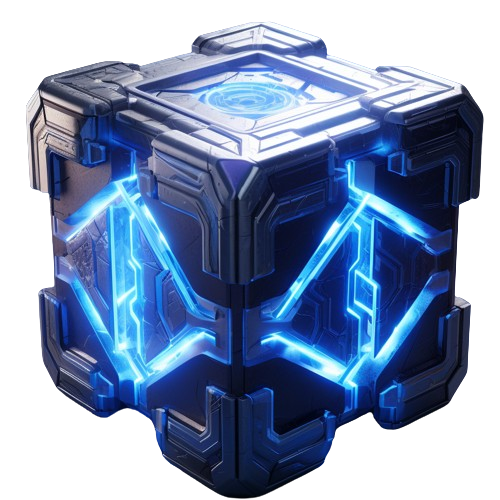}: A Multimodal Spatio-Temporal General Intelligence Model with Ten Modalities via Language as a Reference Framework}

\author{Run Shao, Cheng Yang, Qiujun Li, LinRui Xu, Xiang Yang, Xian Li, MengYao Li, Qing Zhu, Yongjun Zhang, YanSheng Li, Yu Liu, Yong Tang, Dapeng Liu, Shizhong Yang, Haifeng Li~\IEEEmembership{Member,~IEEE,}

\thanks{The work was supported in part by the Major Program Project of Xiangjiang Laboratory under Grant 22XJ01010, in part by the National Natural Science Foundation of China under Grant 61973047 and Grant 42171458, and in part by using Computing Resources at the High-Performance Computing Platform of Central South University. (Corresponding author: Haifeng Li.)}
\thanks{Run Shao, Cheng Yang, Qiujun Li, LinRui Xu, Xiang Yang, Xian Li, MengYao Li and Haifeng Li are with the School of Geosciences and Info-Physics, Central South University, Changsha 410083, China, and also with the Xiangjiang Laboratory, Changsha 410205, China.}

\thanks{Qing Zhu is with the Faculty of Geosciences and Environmental Engineering, Southwest Jiaotong University, Chengdu 611756, China.}

\thanks{Yongjun Zhang and YanSheng Li are with School of Remote Sensing and Information Engineering, Wuhan University, Wuhan 430079, China.}

\thanks{Yu Liu is with the School of Earth and Space Sciences, Peking University, Beijing 100871, China.}

\thanks{Yong Tang, Dapeng Liu are Huawei Technologies Co., Ltd, China.}

\thanks{Shizhong Yang is with BDS Micro Chip Inc, Changsha 410071, China.}

}

\markboth{Journal of \LaTeX\ Class Files,~Vol.~14, No.~8, August~2021}%
{Shell \MakeLowercase{\textit{et al.}}: A Sample Article Using IEEEtran.cls for IEEE Journals}


\maketitle

\begin{abstract}
RGB, multispectral, point and other spatio-temporal modal data fundamentally represent different observational approaches for the same geographic object. Therefore, leveraging multimodal data is an inherent requirement for comprehending geographic objects. However, due to the high heterogeneity in structure and semantics among various spatio-temporal modalities, the joint interpretation of multimodal spatio-temporal data has long been an extremely challenging problem. The primary challenge resides in striking a trade-off between the cohesion and autonomy of diverse modalities. This trade-off becomes progressively nonlinear as the number of modalities expands. Inspired by the human cognitive system and linguistic philosophy, where perceptual signals from the five senses converge into language, we introduce the \textbf{L}anguage \textbf{a}s \textbf{R}eference \textbf{F}ramework (LaRF), a fundamental principle for constructing a multimodal unified model. Building upon this, we propose AllSpark, a multimodal spatio-temporal general artificial intelligence model. Our model integrates ten different modalities into a unified framework, including one-dimensional (language, code, table), two-dimensional (RGB, SAR, multispectral, hyperspectral, graph, trajectory), and three-dimensional (point cloud) modalities. To achieve modal cohesion, AllSpark introduces a modal bridge and multimodal large language model (LLM) to map diverse modal features into the language feature space. To maintain modality autonomy, AllSpark uses modality-specific encoders to extract the tokens of various spatio-temporal modalities. Finally, observing a gap between the model's interpretability and downstream tasks, we designed modality-specific prompts and task heads, enhancing the model's generalization capability across specific tasks. Experiments indicate that the incorporation of language enables AllSpark to excel in few-shot classification tasks for RGB and point cloud modalities without additional training, surpassing baseline performance by up to 41.82\%. Additionally, AllSpark, despite lacking expert knowledge in most spatio-temporal modalities and utilizing a unified structure, demonstrates strong adaptability across ten modalities. LaRF and AllSpark contribute to the shift in the research paradigm in spatio-temporal intelligence, transitioning from a modality-specific and task-specific paradigm to a general paradigm. The source code is available at https://github.com/GeoX-Lab/AllSpark.

\end{abstract}

\begin{IEEEkeywords}
Spatio-temporal Data, Multimodal Machine Learning, Large Language Model, General Intelligence Model.
\end{IEEEkeywords}

\section{Introduction}
\label{section1}

\begin{table*}[h]
\centering
\caption{AllSpark integrates ten spatio-temporal modalities}
\label{tab:support_modalities}
\resizebox{0.9\textwidth}{!}{%
\begin{tabular}{c|ccc|cccccc|c}
\hline
\multirow{2}{*}{Model} &
  \multicolumn{3}{c|}{1D} &
  \multicolumn{6}{c|}{2D} &
  3D \\ \cline{2-11} 
 &
  Language &
  Code &
  Table &
  RGB &
  MSI &
  HSI &
  SAR &
  Traj &
  Graph &
  Point \\ \hline
EarthGPT\cite{zhang2024earthgpt} & \faCheck &  &  & \faCheck &  &  & \faCheck &  &  &  \\
SkySense\cite{guo2024skysense} &  &  &  & \faCheck & \faCheck &  & \faCheck &  &  &  \\
PointLLM\cite{xu2023pointllm} & \faCheck &  &  &  &  &  &  &  &  & \faCheck \\
Table-LLaVA\cite{chen-etal-2023-tablevlm} & \faCheck &  & \faCheck &  &  &  &  &  &  &  \\
OneLLM\cite{han2023onellm} & \faCheck &  &  & \faCheck &  &  &  &  &  & \faCheck \\
Meta-transformer\cite{zhang2023meta} & \faCheck &  & \faCheck & \faCheck &  & \faCheck &  &  & \faCheck & \faCheck \\ \hline
AllSpark & \faCheck & \faCheck & \faCheck & \faCheck & \faCheck & \faCheck & \faCheck & \faCheck & \faCheck & \faCheck \\ \hline
\end{tabular}%
}
\end{table*}

\begin{figure*}[h]
    \centering
    \includegraphics[width=0.75\textwidth]{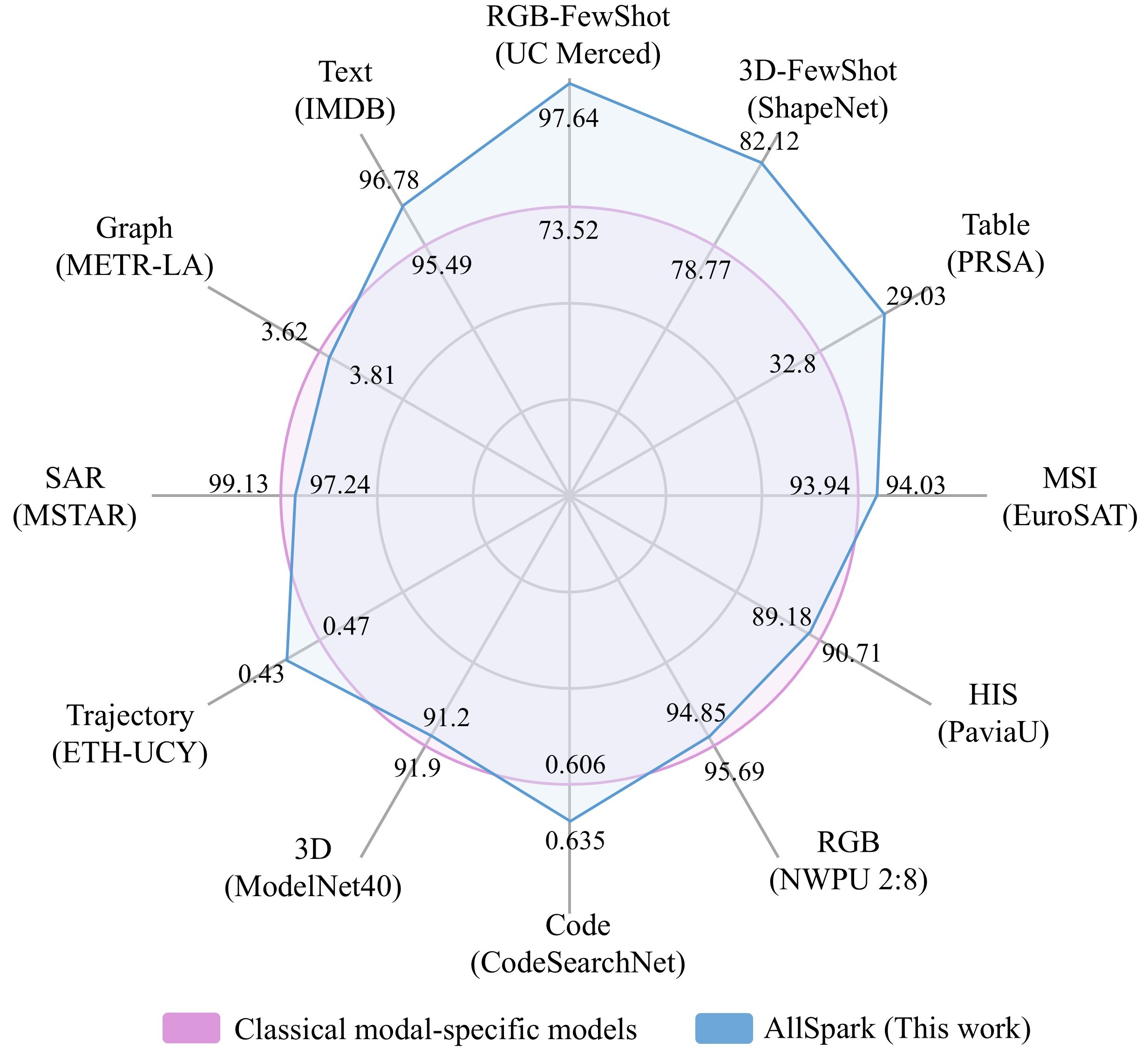}
    \caption{AllSpark demonstrates excellent adaptability across up to 10 heterogeneous modalities and shows outstanding few-shot learning capabilities in RGB and point cloud modalities.}
    \label{fig:radar}
\end{figure*}

\IEEEPARstart{B}{enefiting} from the increasingly diverse observational methods available for spatio-temporal scenes, geographic objects can be described by various spatio-temporal modalities, such as RGB, SAR, multispectral, graph, point cloud and trajectory data \cite{he2023STGC-GNNs,luo2024lsttn,li2024}. Each modality provides unique information about different aspects of the geographic object. Analogous to the human process of perceiving and understanding the world through multiple modalities, such as vision, hearing, and touch, joint interpretation of multimodal data is an inherent requirement for intelligent models to achieve cognition of geographic objects.

However, due to the inherent differences in the mechanisms of each modality, various modalities are often highly heterogeneous in both structure and semantics. For example, in terms of structure, a table is composed of rows and columns. Point clouds are represented by three-dimensional coordinates along with various feature values. A text is composed of sequences of words. In terms of semantics, RGB imagery reflects the electromagnetic characteristics of visible light bands emitted and reflected by geographic objects, whereas SAR imagery reflects the electromagnetic characteristics of microwaves emitted actively by radar for scattering from geographic objects.

For a long time, constrained by the high heterogeneity across various modalities mentioned above, researchers have often developed specific methods based on prior assumptions related to a particular modality or designed multimodal approaches for a few low-heterogeneity modalities. For instance, in single-modal research, C. R. Qi et al. proposed PointNet\cite{qi2017pointnet} for the point cloud modality, emphasizing the invariance of point cloud data ordering and the significance of global and local features. For language modality, Vaswani et al. introduced the transformer\cite{vaswani2017attention}, which focuses on the long-range dependencies within word sequences. For the graph modality, Kipf and Welling proposed the GCN\cite{kipf2016semi,li2021cgnn} based on the adjacency relationships between nodes in a graph. In multimodal research, the fusion of optical and SAR imagery has been widely explored in both traditional and deep learning remote sensing\cite{kulkarni2020pixel}. Moreover, visual-language models have undergone rapid development in recent years\cite{radford2021learning,lu2019vilbert,kim2021vilt,alayrac2022flamingo,li2022blip}. The diverse prior assumptions associated with each modality have resulted in significant gaps between methods designed for different modalities, making it challenging to perceive and understand different modalities using a unified model.

We believe that the key challenge in addressing this issue lies in striking a trade-off between the cohesion and autonomy of diverse modalities. In our paper, "cohesion" refers to the presence of mutually correlated shared information among modalities. For instance, both RGB and SAR images may describe the contour of the same object. "Autonomy," on the other hand, refers to the existence of unique information specific to each modality relative to others. For example, the RGB modality can describe an object's color and texture, while the SAR modality can capture the object's scattering properties in relation to radar waves. Cohesion forms the foundation for the interrelation between modalities, while autonomy highlights the value of multimodal joint interpretation—gaining a complete understanding of an object by integrating multiple modalities.

If we merely project data from different modalities into a shared representation space to emphasize inter-modality cohesion, this approach risks losing modality-specific information, ultimately undermining the unique contributions of each modality and weakening the core value of multimodal collaboration. In contrast, if we excessively stress the autonomy between modalities, it may hinder the establishment of connections among them, limiting the model's ability to simultaneously perceive multiple modalities. Moreover, as the number of modalities increases, balancing cohesion and autonomy becomes progressively more challenging nonlinearly.

We observe that in the process of comprehending the world, humans integrate information from multiple modalities, such as hearing, touch, smell, and vision. The concepts formed through the parsing of these modalities ultimately converge in language. Humans engage in associating, reasoning, and expressive behaviours through language. In other words, language precisely encodes human perception and understanding of the world, providing clear definitions and meanings to abstract concepts from each modality. Inspired by this, we propose the \textbf{L}anguage \textbf{a}s \textbf{R}eference \textbf{F}ramework (LaRF) as a fundamental principle for constructing multimodal models. It means that the abstract concepts derived from each modality should align with language, enabling joint interpretation in the unified representation space of language.

Building upon this, we propose a multimodal spatio-temporal general intelligence model \cite{Tao2023TOV}, AllSpark, that integrates ten different modalities into a unified framework, including one-dimensional (language, code, table), two-dimensional (RGB, SAR, multispectral, hyperspectral, graph, trajectory), and three-dimensional (point cloud) modalities. As shown in Table \ref{tab:support_modalities} and Figure \ref{fig:radar}, previous work either overlooked some important spatio-temporal modalities, such as hyperspectral and trajectory, or focused solely on natural images without considering remote sensing imagery. AllSpark covers a broader range of spatio-temporal modalities, such as multispectral, hyperspectral, graph, trajectory, and more, while demonstrating excellent few-shot learning capabilities and modality adaptability.

To achieve modal cohesion, AllSpark uniformly maps diverse modal features to the language feature space. To maintain the autonomy between modalities, AllSpark introduces specific modal encoders for each modality to extract independent tokens. Given the high heterogeneity among modality data and modality encoders, a significant dimensional gap exists between the tokens of each modality and the language modality. To address this issue, we introduce a modality bridge, a mechanism from perceiver\cite{jaegle2021perceiver}, to accomplish dimensional mapping from each modality's tokens to the language tokens \cite{shao2024}.

Finally, considering the existing gap between the interpretability of the multimodal large language model and the specific downstream tasks, we design task heads and modality-specific text prompts for each downstream task to enhance the model's generalization capability. Given the powerful interpretability capabilities of the multimodal large language model, we adhere to a lightweight design principle in task heads.

Experiments demonstrate that AllSpark, despite lacking expert knowledge in most spatio-temporal modalities and utilizing a unified structure, achieves competitive accuracy in modalities such as RGB and spatio-temporal trajectories compared to state-of-the-art models. Specifically, in the RGB modality, the accuracy of AllSpark is only 0.84 lower than that of the SOTA model, and in the trajectory modality, the average displacement error (ADE) metric differs by only 0.07 compared to that of the SOTA model. Additionally, AllSpark exhibits excellent adaptability in various other modalities, including point cloud, multispectral, hyperspectral, table, graph, and code. Theoretically, our proposed model has the potential for seamless extension to an arbitrary number of modalities.

In other words, our contributions can be summarized as follows:

\begin{itemize}
    \item We first propose a unified multimodal spatio-temporal general model, AllSpark, that successfully integrates ten spatio-temporal modalities into a single model.
    \item Inspired by the human cognitive system and linguistic philosophy, we propose the language as reference framework (LaRF), which offers a novel solution to balance cohesion and autonomy among multiple modalities.
    \item Experiments indicate that AllSpark demonstrates strong few-shot learning capabilities and supports ten modalities. Theoretically, our proposed model has the potential for seamless extension to an arbitrary number of modalities.
\end{itemize}

\section{Related Work}
\label{section2}

Leveraging multimodal data is an inherent requirement for achieving cognitive recognition of geospatial objects. An ideal multimodal model should possess the capability to integrate all the modalities for joint interpretation. Hence, a crucial trend in the research of intelligent methods in the spatio-temporal domain is the continual increase in the number of modalities available for joint interpretation.

Initially, early researchers often constructed single-modal expert models based on prior assumptions about a specific modality, achieving remarkable success within each respective modality. In recent years, with a deeper understanding of single-modal interpretation methods, numerous researchers have attempted to integrate several low-heterogeneity modalities to construct multimodal interpretation approaches. However, as the number of modalities increases, the challenge of balancing cohesion and autonomy among the modalities becomes increasingly difficult. 

In the following, we recall the development of intelligent methods in the spatio-temporal domain from the perspective of the continually increasing number of modalities, and finally, we present the principles and approach of our proposed model, AllSpark.

\subsection{Single-Modal Model}
\label{section2.1}

For one-dimensional modalities, we focus on code, language and table. Given the excellent characteristics of code, such as strict syntax, unambiguous nature, and ability to interact with machines, code is treated as a separate modality. 

Z. Feng et al. pretrained a model, CodeBERT\cite{feng2020codebert}, which facilitates the mutual transfer of information between code and natural language modalities. For the language modality, landmark contributions include a transformer\cite{vaswani2017attention}, BERT\cite{kenton2019bert}, and the GPT series \cite{radford2018improving,radford2019language,brown2020language,achiam2023gpt}, which have inspired subsequent series of works.

The table is one of the commonly used modalities for recording and expressing information. TabNet\cite{arik2021tabnet}, proposed by S. Ö. Arik et al., employs a sequence attention mechanism to achieve feature selection in the Table modality, thereby enabling interpretable and more efficient learning.

We categorize RGB, multispectral, hyperspectral, SAR, graph, trajectory as two-dimensional (2D) modalities.

Among the two-dimensional modalities, standard three-channel RGB images are among the most common. For the RGB modality, the ResNet\cite{he2016deep} proposed by K. He et al., which is based on the importance of visual global and local information, and the vision transformer (ViT)\cite{dosovitskiy2020image} introduced by A. Dosovitskiy et al., which leverages a global attention mechanism, represent two landmark contributions.

An increase in the number of channels in images leads to multispectral and hyperspectral modalities. B. Huang et al. proposed the STDCNN\cite{huang2018urban}, leveraging the characteristic of a greater number of bands in multispectral images to simultaneously model the global spatial and spectral properties of multispectral images. In comparison to multispectral images, hyperspectral images have even more bands, often reaching hundreds. Based on this, X. Yang et al. introduced the R-3D-CNN\cite{yang2018hyperspectral} to further enhance the extraction of spectral features.

In the case of SAR images formed by active microwave radar, S. Chen et al. introduced AConvNet\cite{chen2016target}, a widely used fully convolutional neural network for intelligent SAR image interpretation.

The trajectory modality reflects the temporal changes in the spatial positions of objects. A. Gupta et al. proposed the social generative adversarial networks (GAN)\cite{gupta2018social} based on the characteristic of trajectory multiplicity. This model combines historical trajectory information with social context information to predict multiple plausible future outcomes.

For the graph modality, Kipf and Welling introduced the classic graph convolutional network (GCN)\cite{kipf2016semi}, which is based on the adjacency relationships between nodes in the graph. P. Veličković et al. proposed the GAT\cite{velivckovic2017graph}, which incorporates attention mechanisms into the graph modality.

Finally, we turn our attention to three-dimensional modalities: point cloud.

The point cloud modality captures information about the position, shape, colour, texture, and other aspects of three-dimensional objects. C. R. Qi et al. introduced the classic PointNet\cite{qi2017pointnet} for the point cloud modality, emphasizing the importance of invariance through point data permutation and the significance of global and local features. W. Wu et al. extended convolution operations to three-dimensional point clouds with the introduction of PointConv\cite{wu2019pointconv}.

\subsection{Multimodal Model}
\label{section2.2}

While the single-modal methods in Section 2.1 have demonstrated excellent performance within their respective modalities, they often face challenges in generalizing across multiple modalities due to their construction based on specific prior assumptions. Recognizing the intrinsic requirement for intelligent models to utilize multimodal data for geographic object cognition, numerous researchers have endeavoured to balance the cohesion and autonomy among modalities to construct multimodal models.

A. Sadeghian et al. extended the social GANs\cite{gupta2018social} and introduced RGB images to enhance scene data in Sophie\cite{sadeghian2019sophie}, achieving better results in trajectory prediction tasks. Recognizing the high complementarity between RGB and SAR images, L. H. Hughes et al. proposed a three-step deep neural network framework that utilizes a universal prediction of matching regions, generates heatmaps, and eliminates outliers to match RGB and SAR images\cite{hughes2020deep}. X. Li et al. introduced the DTCDN\cite{li2021deep}, a model that employs a GAN network to migrate RGB and SAR images to the same feature space, facilitating target detection. J. Yang et al. proposed a dual-stream convolutional network that uses high-resolution multispectral images to enhance the spatial resolution of hyperspectral images\cite{yang2018hyperspectralandmultispectral}.

To achieve joint interpretation of multimodal data, traditional multimodal models typically design specific architecture based on the priors of certain modalities. For example, Hang et al. proposed Coupled CNNs \cite{hang2020classification}, which consist of two CNN networks to extract spectral-spatial features from hyperspectral data and elevation information from LiDAR data. They ultimately use both feature-level and decision-level fusion methods to integrate the heterogeneous features of the two modalities. Similarly, Zhang et al. proposed SLA-Net \cite{zhang2023morphological}, which designs specific network structures to extract and fuse spatial information and morphological characteristics from hyperspectral imagery. These methods are typically designed for specific purposes and modalities, making it difficult to extend them to more modalities.

Additionally, Gao et al. proposed DFINet \cite{gao2021hyperspectral}, which extracts self-correlation and cross-correlation between multimodal data to deeply fuse hyperspectral and multispectral modality features. Likewise, Hong et al. introduced S2FL \cite{hong2021multimodal}, which extracts Modality-Specific Subspaces for each modality and a Shared Subspace for all modalities, finally obtaining multimodal interpretation results through a unified projection. The issue with such methods is the lack of a unified alignment reference, limiting them to collaboration between a few modalities. As the number of modalities increases, the complexity of aligning multiple modalities will grow exponentially.

Notably, CLIP\cite{radford2021learning}, proposed by A. Radford et al., associates the RGB modality with the text modality using contrastive learning \cite{zzy2023,pengjian}. With pretraining guided by weak supervisory signals from text, CLIP has demonstrated outstanding capabilities in both visual single-modal tasks and visual-language multimodal tasks, inspiring a series of subsequent works\cite{lu2019vilbert,kim2021vilt,alayrac2022flamingo,li2022blip}. Y. Zhang et al. introduced a meta-transformer\cite{zhang2023meta}, leveraging the contrastive learning paradigm from CLIP to pretrain a universal backbone network under the visual-language modality. It exhibits multimodal generalization abilities across various modalities, such as point cloud, infrared, and hyperspectral data. J. Han et al. directly employed a multimodal large language model as a universal backbone network, proposing the One-LLM, which successfully unifies eight modalities, namely, images, audio, videos, and points\cite{han2023onellm}. The success of these approaches implies the unique role of language modalities in multimodal models.

Building upon the aforementioned efforts, we systematically propose the fundamental principle of the LaRF. Guided by this principle, we balance cohesion and autonomy among diverse modalities and introduce a general intelligent model named AllSpark, which unifies ten spatio-temporal modalities and possesses the potential to extend to an arbitrary number of modalities.

\section{Method}
\label{section3}

\subsection{Language as Reference Framework}
\label{section3.1}

We observe that in the process of comprehending the world, humans integrate information from multiple modalities, such as hearing, touch, smell, and vision. The concepts formed through the parsing of these modalities ultimately converge in language. Humans engage in associating, reasoning, and expressive behaviours through language. In other words, language precisely encodes human perception and understanding of the world, providing clear definitions and meanings to abstract concepts from each modality.

Inspired by our observation, we introduce the fundamental principle of \textbf{L}anguage \textbf{a}s \textbf{R}eference \textbf{F}ramework (LaRF) to balance cohesion and autonomy among multiple modalities.

\begin{figure}[!t]
    \centering
    \includegraphics[width=3.5in]{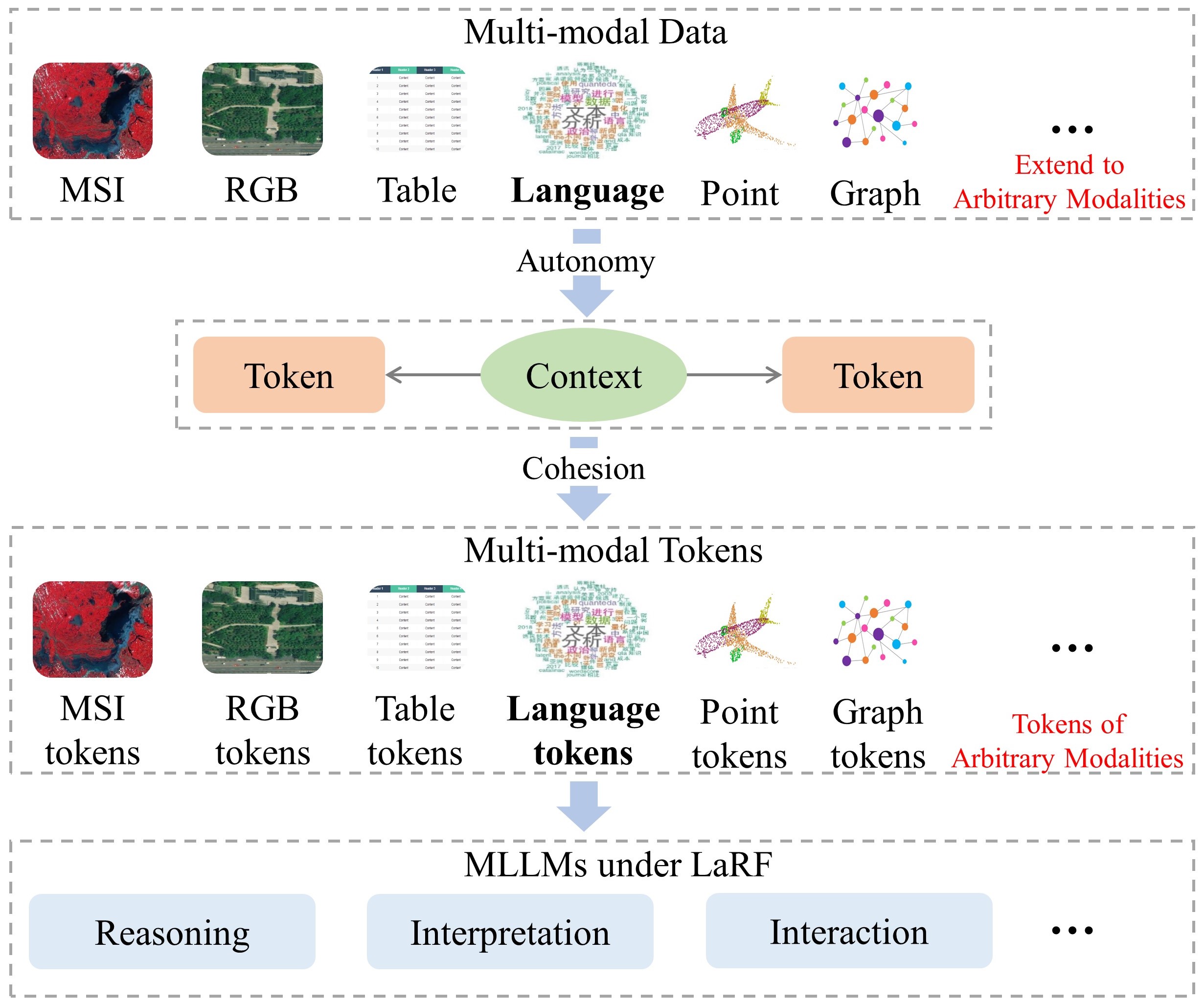}
    \caption{Guided by the LaRF principle, multimodal data are transformed into a token-context structure akin to language, based on their respective prior assumptions. This approach preserves the autonomy of each modality while achieving cohesion between them, enabling the interpretation of multimodal data within a unified language representation space.}
    \label{fig:tokenRepresentation}
\end{figure}

In terms of the cohesion of multimodalities, the high heterogeneity between multiple spatio-temporal modalities is a major challenge, while the LaRF principle defines the alignment anchor between multimodalities as language explicitly. As shown in Figure \ref{fig:tokenRepresentation}, we observe that language is encoded by tokens and their contexts, and this structure can be extended to most spatio-temporal modalities. Therefore, we can align highly heterogeneous spatio-temporal modalities to language modalities in structure and semantics, enabling multimodal interpretation in the unified representation space of language. Additionally, the pivotal role of natural language prompts is a key factor in the LaRF principle’s ability to achieve cohesion across modalities.

In contrast, we can independently encode multiple spatio-temporal modalities into token sequences under their respective prior assumptions, so the LaRF principle does not lead to the loss of modal autonomy. More importantly, LaRF is not dependent on specific modalities; therefore, theoretically, as long as token representations of modalities can be obtained, the multimodal model guided by LaRF can be extended to arbitrary modalities.

In summary, the significance of LaRF is as follows:

\begin{enumerate}
    \item \textbf{Alignment Capability}: Language can accurately encode both cohesion and autonomy information across multiple modalities. Aligning each modality with the language modality enables a unified representation in the same feature space, addressing the challenge of high heterogeneity among modalities.
    \item \textbf{Reasoning Capability}: Language, as a tool for human thought and expression, inherently possesses the ability to perform complex reasoning. Each modality, when represented in a unified space with LaRF, inherits the reasoning capability of language, unlocking the potential for multimodal joint reasoning.
    \item \textbf{Interpretability}: Deep learning methods have long been characterized as "black boxes." However, a multimodal intelligent system constructed based on the LaRF can directly leverage language as a tool. This facilitates the output of interpretable reasoning chains that humans can understand, thereby achieving true explainable artificial intelligence.
    \item \textbf{Interactivity}: Language not only aids humans in understanding intelligent models but also facilitates intelligent models in understanding humans. In an intelligent system guided by the LaRF, humans can directly express their needs using natural language. This iterative correction of the model's output based on human interaction will become a new paradigm for the training and inference of intelligent models.
    \item \textbf{Scalability}: The multimodal system guided by the LaRF is agnostic to specific modalities. New modalities need to establish a mapping to the language model only to participate in joint reasoning with other modalities. Therefore, theoretically, a multimodal model based on LaRF can be extended to an arbitrary number of modalities.
\end{enumerate}

\begin{figure*}[!t]
    \centering
    \includegraphics[width=7in]{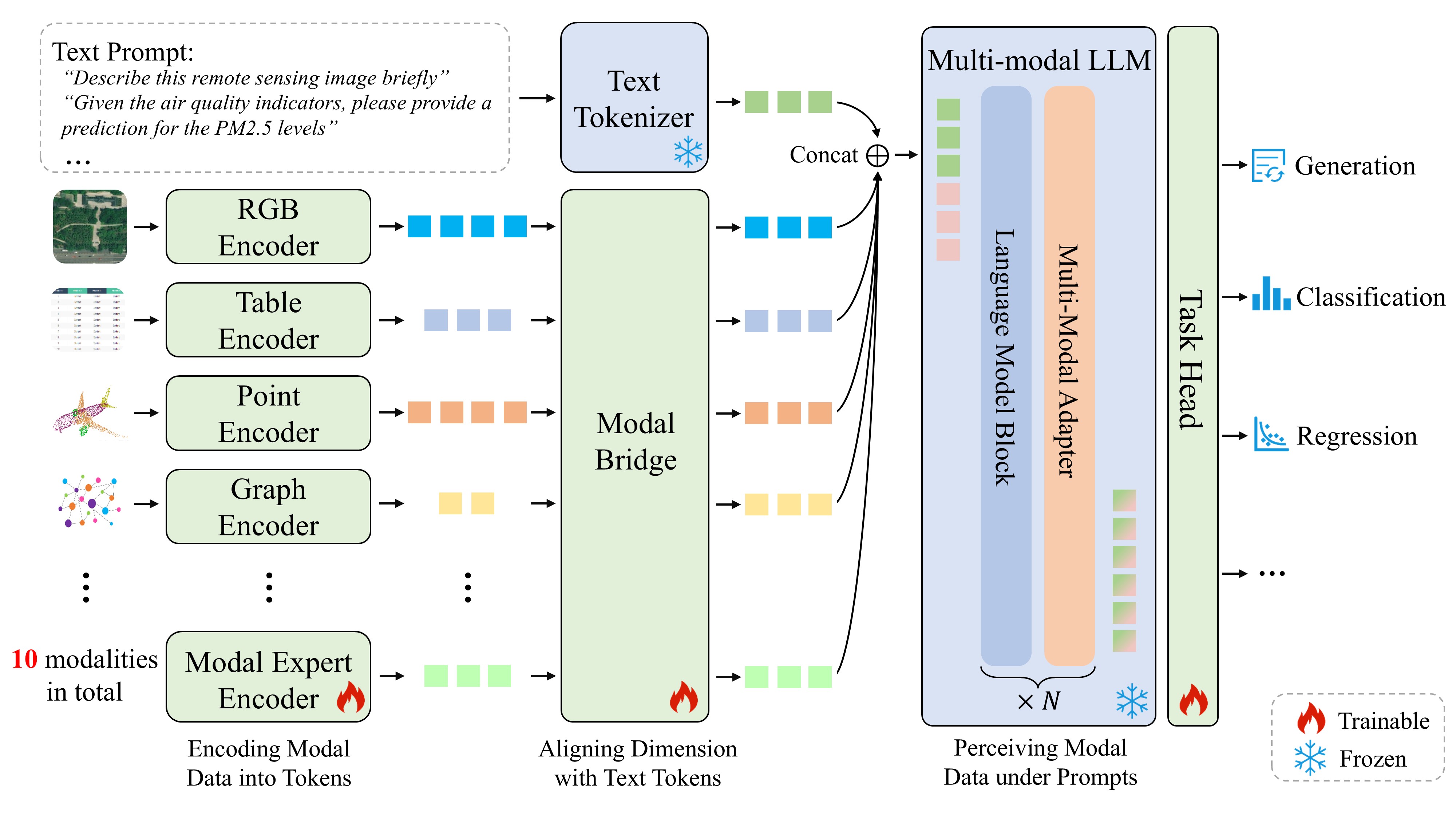}
    \caption{AllSpark Architecture. Multimodal data are extracted by their respective modal encoders into token sequences. Following dimension alignment with modality-specific text prompt tokens via a modal bridge, both the text prompt tokens and modality tokens are passed into a large language multimodal model for interpretation. The interpretation results are then aligned with downstream tasks through task-specific heads.}
    \label{fig:Model}
\end{figure*}

\subsection{Overview}
\label{section3.2}

\begin{table*}[!t]
  \centering
  \caption{List of symbols}
    \begin{tabular}{p{5em}p{15em}|ll}
    \toprule
    \multicolumn{2}{c|}{\textbf{Module}} & \multicolumn{2}{c}{\textbf{Variants}} \\
    \midrule
    Symbol & Meaning & \multicolumn{1}{p{5em}}{Symbol} & \multicolumn{1}{p{15em}}{Meaning} \\
    \midrule
    $M$     & AllSpark & \multicolumn{1}{p{5em}}{ $m_i$ } & \multicolumn{1}{p{15em}}{modal i} \\
    $f_i$ & modal encoder for $m_i$ & $q$ & query vectors of the bridge \\
    $\Phi$ & modal bridge & $s_i$ & tokens of $m_i$ \\
    $T$ & text tokenizer & $p_i$ & text prompt for $m_i$ \\
    $F$ & Multimodal LLM & $\theta$ & model parameters \\
    $H_{task}$ & Task Head & $y$ & label \\
    $L$ & loss function & $t_i$ & tokens of $m_i$ \\
    ${Enc}_n$ & n-layer transformer encoders & $W$ & weights of the linear layer \\
    $Emb$   & embedding layer & $w_i$ & word of the code or text \\
    ${Enc}_{ResNet}$ & ResNet &       &  \\
    $Grouper$ & point grouper of the PointBERT\cite{yu2022point} &       &  \\
    $Conv1d$ & 1D convolution layer &       &  \\
    $FFN$   & feedforward network &       &  \\
    $\sigma$ & softmax &       &  \\
    \bottomrule
    \end{tabular}%
  \label{tab:symbols}%
\end{table*}

Guided by the principles mentioned above, AllSpark consists of five modules: the modal encoder, modal bridge, text tokenizer, multimodal large language model, and task head. The overall architecture is depicted in Figure \ref{fig:Model}.

To maintain autonomy among modalities, we designed modality-specific encoders to encode highly heterogeneous data into modality-independent tokens (for details, see Section \ref{section3.3}). However, the dimensions of tokens outputted by modal encoders are still inconsistent. To parse in the unified representational space of language, we introduced the modal bridge from the Lynx\cite{zeng2023matters}. The modal bridge aims to project tokens from each modality into the dimension of the multimodal large language model (Section \ref{section3.4} for details). The formalization of this process is as follows:

\begin{equation}
s_i=\Phi\left(f_i\left(m_i\right),q\right)
\end{equation}

Here, $\{m_{RGB},m_{MSI},m_{HSI},\ldots,m_i,\ldots\}$ represents inputs from various modalities, where $m_{RGB}$ represents the RGB modality, $m_{MSI}$ represents the multispectral modality, and so on. $f_i$ is defined as the modal encoder for the $m_i$ modality, $\Phi$ represents the modal bridge, and $q \in R^{N * D}$represents $N$ learnable vectors of dimension $D$ in the modality bridge, where $D$ is set to 4096, representing the dimensionality of the multimodal large language model. The input data $m_i$ of each modality are mapped to a token sequence $s_i \in R^{N * D}$ of the same dimension as the language model. Table \ref{tab:symbols} summarizes the main mathematical symbols and their meanings in the model.

To achieve cohesion among modalities, we employ a unified multimodal large language model to parse data from various modalities. The text tokenizer and multimodal large language model in AllSpark are based on the visual-language model Lynx\cite{zeng2023matters}. To extend Lynx to ten spatio-temporal modalities, we designed specific text prompts for each modality to guide the model in correctly parsing information from each modality. Additionally, Lynx incorporates several lightweight multimodal adapter layers internally to accommodate multimodal inputs. We continue this design and do not freeze the parameters of the adapter layers during training to enhance the model's adaptability to other spatio-temporal modalities. Finally, we acknowledge that a gap exists between the parsing results of the model and those of the downstream task. Therefore, we design specific task heads for each task to enhance the model's generalization capability.

The entire model M can be formalized as:

\begin{equation}
M\left(m_i,p_i\right)=H_{task}\left(F\left(s_i\oplus T\left(p_i\right)\right)\right)
\end{equation}

where $p_i$ represents the text prompt of $m_i$, $T$ denotes the text tokenizer, $\oplus$ indicates the concatenation operation of text tokens and modality tokens in the sequence, $F$ represents the multimodal large language model, and $H_{task}$ signifies the task head.

All the tasks in our experiments are supervised tasks, with $y$ denoting the labels, $L$ representing the loss function, and $\theta$ representing the learnable parameters in our model. The optimization objective of the model can be formalized as follows:

\begin{equation}
\theta_i=\mathop{\arg\min}_{\theta} {L\left(y,M\left(m_i,p_i\right);\theta_i\right)} 
\end{equation}

\subsection{Independent Encoder for Each Modality}
\label{section3.3}

The modal encoder aims to encode the raw data of each modality into a token sequence, formalized as $t_i=f_i\left(m_i\right)$, where $t_i \in R^{n * d}$. We designed different modal encoders for each modality to maintain autonomy among modalities. The following provides individual introductions for each modality:

\subsubsection{1D modal}
\label{section3.3.1}
\ 
\newline
\indent Code \& Language: Code is essentially a specialized form of language. But due to its distinct properties such as having a strict syntax, being unambiguous, and being capable of interacting with machines, we separate it as a distinct modality. Therefore, to avoid ambiguity, we will use "text" and "language" interchangeably to distinguish between the natural language and code modalities. Exploring intelligent methods for the code modality is crucial for reliable AI reasoning and achieving interaction between intelligent models and the real world. Since the Lynx model is a language model, we do not design an additional modal encoder for the code and text modalities. Instead, we directly utilize Lynx's text tokenizer, i.e., $f\left(m_{Text/Code}\right)=T\left(m_{Text/Code}\right)$, where $m_{Text/Code}=\{w_1,w_2,w_3,\ldots\}$ represents the sequence of words in the text or code.

Table: A table can be viewed as a sequence of rows, with each row containing several fields, or columns, i.e., $m_{Table} \in R^{ {row} * {col}}$. The modal encoder for the table modality inherits the design from TabFormer\cite{padhi2021tabular}: first, based on the different degrees of discreteness for each column attribute, we use independent embedding layers to encode discrete and continuous values separately. Subsequently, we employ a single-layer transformer encoder to further extract features. The entire process can be formalized as $f\left(m_{Table}\right)={Enc}_1\left(Emb(m_{Table})\right)$.

\subsubsection{2D modal}
\label{section3.3.2}
\ 
\newline
\indent RGB: RGB imagery represents the visible light spectrum and reflects the electromagnetic characteristics of objects that emit or reflect visible light waves. It is the most common modality in the field of computer vision. RGB imagery is a standard three-band image, i.e., $m_{RGB} \in R^{H * W * 3}$. For the modal encoder of this modality, we adopted the visual encoder from the Lynx model: EVA\cite{fang2023eva}. The EVA is a large visual model composed of 40 stacked transformer blocks with a width of 1408. During the experiments, AllSpark loaded the official weights of the EVA model and froze them during training.

MSI: Multispectral imagery is a modality extensively studied in the remote sensing field. It incorporate multiple nonvisible light bands, such as near-infrared, shortwave infrared, coastal atmospheric aerosol, and cirrus bands. Therefore, the number of channels in MSIs is usually greater than the three bands in RGB imagery, i.e., $m_{MSI} \in R^{H * W * C}$, where $C>3$. We extended the PatchEmbed of the standard ViT\cite{dosovitskiy2020image}, modifying its channel count to match the number of bands in the input multispectral imagery. This adaptation allows it to serve as the feature encoder for the multispectral modality.

HSI: Hyperspectral imagery increases the number of bands compared to multispectral imagery, often reaching hundreds of bands, with each band containing rich information. Unlike RGB and multispectral imagery, where an image serves as a single sample, in hyperspectral imagery, all bands for each pixel are treated as a single sample, i.e., $m_{HSI} \in R^{1 * 1 * C}$. In the modal encoder for hyperspectral imagery, we first use a linear projection layer to expand the feature dimensions for each pixel. This process is formalized as $W * m_{HSI}^T$, where $W \in R^{1 * d}$ is the weight matrix of the linear projection layer. Subsequently, we use a 12-layer standard transformer encoder to extract its features. The entire process can be represented as $f\left(m_{HSI}\right)={Enc}_{12}\left(W * m_{HSI}^T \right)$.

Trajectory: The trajectory modality reflects the changing information of an object over time and space and is composed of a series of two-dimensional coordinate points, i.e., $m_{Trajectory} \in R^{l * 2}$, where $l$ represents the sequence length of trajectory points. The encoder for the trajectory modality inherits the design from TUTR\cite{shi2023trajectory}: first, a linear layer is used to expand the dimensions of the two-dimensional trajectory features. This step can be formalized as ${W * m}_{Trajectory}$, where $W \in R^{d * \left(l * 2\right)}$ is the weight matrix of the linear projection layer. Then, we use a 2-layer transformer encoder to extract its features. The entire process can be formalized as $f\left(m_{Trajectory}\right)={Enc}_2\left({W * m}_{Trajectory}\right)$.

SAR: The synthetic aperture radar (SAR) modality is a type of active remote sensing that reflects the electromagnetic characteristics of objects with respect to microwave backscatter. Due to differences in polarization modes, the final product of SAR imagery is typically a two-band or single-band image, i.e., $m_{SAR}\in R^{H * W * 2}$. Therefore, we designed a simple three-layer convolutional network as the modal encoder.

Graph: A graph is constructed from a series of nodes and edges, where the attributes of the nodes and the adjacency attributes of the nodes reflect the majority of the features of the graph, i.e., $m_{Graph} \in R^{K * d}$, where K is the number of nodes and d is the node feature dimension. In AllSpark, the modality encoder of a graph is based on the STAEformer\cite{liu2023spatio,zhu2022kst}, whose main design idea is to first use a linear layer to extend its feature dimension and then use several embedding layers to separately encode features such as the node's characteristics, spatial characteristics, and temporal characteristics. The entire process can be formalized as $f\left(m_{Graph}\right)={Emb}_{node}\left({W * m}_{Graph}^T\right)\oplus{Emb}_{spatial}\left({W * m}_{Graph}^T\right)\oplus{Emb}_{time}\left({W * m}_{Graph}^T\right)$, where $W \in R^{{hidden} * d}$ is the weight of the linear layer.

\subsubsection{3D modal}
\label{section3.3.3}
\ 
\newline
\indent Point Cloud: A point cloud is typically composed of three-dimensional coordinates and feature values $m_{PointCloud}\in R^{K*\left(d+3\right)}$, where K represents the number of three-dimensional points and d\ represents the dimensionality of the point cloud features, reflecting information such as the spatial position, shape, colour, and texture of objects. The encoder for the point cloud modality inherits the design from PointBERT\cite{yu2022point}: first, point cloud data are grouped and encoded to unify the number of points simultaneously inputted. This step can be represented as $PointGroup=Grouper\left(m_{PointCloud}\right) \in R^{G*N*3}$, where $G$ represents the number of groups and $N$ represents the number of points in each group. Next, the grouped results are input into a one-dimensional convolutional layer to extract feature vectors for each group:$f_{Group}=Conv1d\left(PointGroup\right)\in R^{G*d}$. Finally, the feature vectors for each group are input into a standard 12-layer transformer encoder to extract their global features. The entire process can be formalized as $f\left(m_{PointCloud}\right)=Enc_{12}\left(Conv1d(Grouper(m_{PointCloud}))\right)$.

\subsection{Modal2Language Bridge}
\label{section3.4}

Although the modal encoders have transformed data from various heterogeneous modalities into a unified token sequence, there are still differences in dimensions between different modal tokens, making it difficult to perceive by a multimodal large language model. The modal bridge, based on the Perceiver\cite{jaegle2021perceiver}, aims to perform dimensional projection from tokens of various modalities to tokens of the language modality. In its implementation, the modal bridge consists of stacked cross-attention layers and feedforward neural network layers.

In the cross-attention layer, we predefine a learnable query vector $Q\in R^{N* D}$, where $D$ is the internal dimensionality of the language model and $N$ serves as a hyperparameter that can be flexibly adjusted to accommodate inputs from different modalities. The keys and values in the cross-attention layer are the features outputted by the modal encoders.

The feedforward neural network inherits the classic design from the original transformer and consists of two linear layers with an inserted activation layer.

The entire process can be formalized as follows:

\begin{equation}
\Phi\left(Q,s_i\right)={FFN}\left (\sigma \left (\frac{Q W_q^T\left ( s_i W_k^T \right )^T }{\sqrt{D}}\right )s_i W_v^T   \right )
\end{equation}

Here, $W_q \in R^{D* {hidden}}$, $W_k \in R^{d* {hidden}}$, and $W_v \in R^{d* {hidden}}$ are the linear projection layer weights defined inside the cross-attention layer for $Q$, $K$ and $V$, respectively. $\sigma$ denotes the softmax operation.

\subsection{Task-Guided Text Prompts and Task Heads}
\label{section3.5}

To extend the visual-language multimodal model to ten spatio-temporal modalities without intervention from modality expert knowledge, we designed specific text prompts and task heads for each modality and task. Text prompts are used to guide the multimodal language model in correctly interpreting each modality's data, while task heads are employed to match the model's parsing results with specific downstream tasks.

We manually designed one to four specific text prompts for each modality. During the training process, to enhance model performance, we employed a strategy of diversifying prompts, randomly selecting one prompt for each forward pass. However, during testing, for the sake of result stability and reproducibility, the prompt was fixed to be the first prompt among all prompts. Table \ref{tab:prompts} provides a list of all the text prompts.

To ensure the transferability of modalities across different tasks, the design principle for task heads is to be as simple and lightweight as possible. 

For classification tasks or downstream tasks that can be formalized as classification tasks, we uniformly use a simple single-layer linear layer as the task head. For instance, we implemented standard classification tasks on the RGB, MSI, SAR, and point cloud modalities. Although the task involving the HSI modality is segmentation, it can be formalized as a per-pixel classification task. Therefore, a single-layer linear layer is used as the task head for the mentioned modalities. 

For regression tasks on the table, trajectory, and graph modalities, we also use a linear layer to perform regression predictions. The only difference from classification tasks is the addition of an unscaled operation without learnable parameters. Since the Lynx itself is a language model, the code and text modalities directly use its native text decoder.

\begin{table}[]
  \centering
  \caption{Summary of hyperparameters}
    \begin{tabular}{c|l|l|l|p{4em}}
    \toprule
    \multicolumn{1}{c|}{Dimension} & Modal & Max Lr & Max Epochs & Warm-up Epochs \\
    \midrule
    \multicolumn{1}{c|}{\multirow{3}[4]{*}{1D}} & Language  & $9.0 \times 10^6$ & 5     & 1 \\
\cmidrule{2-5}          & Code  & $1.0 \times 10^5 $ & 4     & 1 \\
\cmidrule{2-5}          & Table & $2.0 \times 10^5$ & 30    & 3 \\
    \midrule
    \multicolumn{1}{c|}{\multirow{6}[8]{*}{2D}} & RGB   & $5.0 \times 10^5 $ & 50    & 5 \\
\cmidrule{2-5}          & MSI   & $2.0 \times 10^5$ & 50    & 5 \\
\cmidrule{2-5}          & HSI   & $1.0 \times 10^4$ & 30    & 3 \\
\cmidrule{2-5}          & SAR   & $9.0 \times 10^6$ & 30    & 3 \\
\cmidrule{2-5}          & Trajectory & $1.0 \times 10^5$ & 30    & 5 \\
\cmidrule{2-5}          & Graph & $8.0 \times 10^5$ & 10    & 2 \\
    \midrule
    \multicolumn{1}{c|}{3D} & Point Cloud & $3.0 \times 10^5$ & 100   & 10 \\
    \bottomrule
    \end{tabular}%
  \label{tab:hyperparameters}%
\end{table}%

\section{Experiment}
\label{section4}

\subsection{Setup}
\label{section4.1}

Our experiments aim to demonstrate: (1) the distinct advantages of AllSpark over traditional models, and (2) AllSpark's ability to understand 10 spatio-temporal modalities simultaneously.

For the former, we believe that since language as reference framework, AllSpark's advantage over traditional models lies in the richer semantic meaning of its features. In large language models\cite{devlin2018bert, radford2019language, brown2020language, touvron2023llama}, a common way to measure this property is through few-shot learning. Therefore, we evaluated AllSpark's performance on few-shot classification tasks in the RGB and point cloud modalities without any extra training steps. The experimental results can be found in Section \ref{section4.4}.

For the latter, we select a task for each modality to conduct the evaluation. Following the principles of simplicity and reproducibility, we choose the widely studied datasets in each modality's respective field and employ similar experimental settings across all modalities. Specifically, we use the AdamW optimizer with a learning rate schedule based on cosine annealing. The hyperparameters are adjusted slightly in terms of training epochs and learning rates for different experiments. The specific details on dataset selection can be found in Section \ref{section4.2}. Table \ref{tab:hyperparameters} summarizes the hyperparameter settings for the experiments on each modality. The experimental results can be found in Section \ref{section4.3}.

\subsection{Dataset}
\label{section4.2}

Below, we will provide detailed explanations of datasets in order:

Language: The IMDB\cite{maas2011learning} dataset is a binary sentiment analysis dataset consisting of 50,000 reviews from the Internet Movie Database (IMDb) labelled as positive or negative. Additionally, the dataset includes some unlabelled data. In our experiments, only the labelled data from the IMDB dataset were utilized for supervised sentiment classification tasks.

Code: CodeSearchNet\cite{husain2019codesearchnet} is a large-scale dataset of function code and its documentation from GitHub that covers six programming languages: Go, Java, JavaScript, PHP, Python, and Ruby. The task performed on the code modality is code document generation, and we tested it on the Ruby and JavaScript.

\begin{table*}[!t]
\caption{The few-shot classification results for the RGB modality}
\label{tab:fs_RGB}
\centering
\resizebox{0.8\textwidth}{!}{
\begin{tabular}{cccccc}
\hline
\multirow{2}{*}{Model} & \multirow{2}{*}{Training-free} & \multicolumn{2}{c}{UC Merced} & \multicolumn{2}{c}{RS19} \\ \cline{3-6} 
             &   & 5way-1shot & 5way-5shot & 5way-1shot & 5way-5shot \\ \hline
MatchingNet\cite{vinyals2016matching}  & \ding{55} & 48.18      & 67.39      & 67.68      & 85.01      \\
RelationNet\cite{sung2018learning}  & \ding{55} & 50.07      & 65.22      & 65.01      & 79.75      \\
ProtoNet\cite{snell2017prototypical}     & \ding{55} & 53.85      & 71.23      & 76.36      & 85.00      \\
DLA-MatchNet\cite{li2020dla} & \ding{55} & 53.76      & 63.01      & 68.27      & 79.89      \\
SPNet\cite{cheng2021spnet} & \ding{55} & 57.64      & 73.52     & 81.06      & 88.04\\ \hline
AllSpark     & \ding{52} & \textbf{95.58} &  \textbf{97.64}      &  \textbf{97.16}      &  \textbf{98.94}      \\ \hline
\end{tabular}
}
\end{table*}

\begin{table*}[!t]
\caption{The few-shot classification results for the point modality}
\label{tab:fs_Point}
\centering
\resizebox{0.8\textwidth}{!}{%
\begin{tabular}{cccccc}
\hline
\multirow{2}{*}{Model} & \multirow{2}{*}{Training-free} & \multicolumn{2}{c}{ShapeNet} & \multicolumn{2}{c}{ScanObject} \\ \cline{3-6} 
            &   & 5way-1shot     & 5way-5shot     & 5way-1shot     & 5way-5shot     \\ \hline
FSLGNN\cite{garcia2017few}      & \ding{55} & 64.98          & 76.14          & 29.91          & 32.77          \\
RelationNet\cite{sung2018learning} & \ding{55} & 65.88          & 76.25          & 45.32          & 55.43          \\
ProtoNet\cite{snell2017prototypical}    & \ding{55} & 65.96          & 78.77          & \textbf{44.75} & \textbf{59.81} \\ \hline
AllSpark    & \ding{52} & \textbf{67.20} & \textbf{82.12} & 35.70          & 53.33          \\ \hline
\end{tabular}%
}
\end{table*}

Table: The PRSA\cite{zhang2017cautionary} dataset is a collection of air quality data from multiple stations in Beijing that contains hourly measurements of air pollutants. The data spans from March 1, 2013, to February 28, 2017, across 12 monitoring stations. In our experiments, we used various features, including time, station information, 4 air pollutant variables (SO2, NO2, CO, and O3), and 6 meteorological variables (temperature, pressure, dew point temperature, amount of precipitation, wind speed, and wind direction). The task was to predict the concentration of PM2.5. We split the data into training (40\%) and testing sets.

RGB: NWPU-RESISC45\cite{cheng2017remote} is a large-scale open dataset for visible light remote sensing image scene classification. The dataset included 45 land use categories, such as airplanes, baseball diamonds, beaches, and commercial areas. Each category included 700 remote sensing images, for a total of 31,500 images. The image size is 256*256 pixels, and we selected a version of the dataset split by the official release using 20\% of the data for training.

MSI: The EuroSAT\cite{helber2019eurosat} dataset is a multispectral dataset for land use and land cover (LULC) classification. The samples are sourced from the Sentinel-2 optical satellite and include all 13 bands. The data are categorized into 10 classes for a total of 27,000 images. We adopted a random 9:1 split for training and testing.

HSI: The Pavia University dataset is a high-spatial-resolution hyperspectral dataset acquired by the ROSIS sensor. It comprises 103 bands with a size of 610*340 pixels. The dataset includes nine land cover categories, such as asphalt, meadows, and gravel. We used a 4:6 split for training and testing.

Trajectory: The ETH-UCY\cite{pellegrini2009you,lerner2007crowds} dataset is a widely used benchmark for pedestrian trajectory prediction and is divided into five subsets: ETH, HOTEL, UNIV, ZARA1, and ZARA2. In our experiments, we utilized the ETH subset.

SAR: The MSTAR\cite{keydel1996mstar} dataset is a synthetic aperture radar dataset designed for military stationary target recognition that comprises ten categories of military targets. We employed the standard operating conditions (SOCs) dataset preprocessing method proposed by S. Chen et al. \cite{chen2016target}, ensuring that the serial numbers and target configurations are consistent between the test and training sets while the aspects and depression angles differ.

Graph: METR-LA is a traffic dataset collected from loop detectors on the Los Angeles highways spanning from March 1, 2012, to June 30, 2012. The task is traffic flow prediction.

Point cloud: ModelNet40\cite{wu20153d} is a synthetic point cloud dataset consisting of 40 object categories and a total of 12,311 point cloud objects. We follow the official dataset split, with 9,843 objects used for training and 2,468 for testing.

\subsection{Few-shot learning}
\label{section4.4}

To demonstrate AllSpark's unique advantages over traditional models, we tested its few-shot performance on the RGB and point cloud modalities. It is worth noting that traditional few-shot learning methods typically require additional training steps. For instance, common approaches like ProtoNet\cite{snell2017prototypical} and MatchingNet\cite{vinyals2016matching} involve randomly splitting support and query sets on the training set for supervised training, a step known as meta-learning. Thanks to the integration of natural language, AllSpark requires no extra training step and can directly evaluate few-shot classification accuracy on the test set, significantly outperforming baseline models.

\subsubsection{RGB}
\label{section4.4.1}

We evaluated AllSpark's RGB image few-shot classification performance on the UC-Merced\cite{yang2010bag} and WHU-RS19\cite{Dai2011WHURS19, Xia2010WHURS19} datasets, following the dataset splits from \cite{li2020dla}. Specifically, the UC-Merced dataset uses six classes—Beach, Golf course, Mobile home park, River, Sparse residential, and Tennis court—as the test set, while the WHU-RS19 dataset uses five classes—Commercial, Meadow, Pond, River, and Viaduct—as the test set. AllSpark does not require meta-learning on the training set and is evaluated on 600 episodes directly on the test set, with 15 query samples per episode. As shown in the Table \ref{tab:fs_RGB}, we report the results for both 5-way 1-shot and 5-way 5-shot settings. The experiments demonstrate that AllSpark significantly outperforms baseline models without requiring any training, highlighting the advantage of LaRF.

\subsubsection{Point}
\label{section4.4.2}

We evaluated AllSpark's point cloud few-shot classification performance on the ShapeNet\cite{chang2015shapenet} and ScanObjectNN\cite{uy-scanobjectnn-iccv19} datasets, following the dataset settings from \cite{ye2023closer}. Similar to the RGB modality, AllSpark does not require meta-learning on the training set and is evaluated on 700 episodes directly on the test set, with 15 query samples per episode. As shown in the Table \ref{tab:fs_Point}, we report the results for both 5-way 1-shot and 5-way 5-shot settings. The experiments demonstrate that AllSpark surpasses most baseline models even on previously unseen point cloud datasets, showcasing its outstanding semantic richness and generalization capabilities.

\subsection{Ability to understand 10 spatio-temporal modalities}
\label{section4.3}

\subsubsection{RGB}
\label{section4.3.1}

We evaluated the performance of AllSpark on the RGB image scene classification task using the NWPU-RESISC45 dataset, with the top-1 accuracy as the evaluation metric. AllSpark leverages expert knowledge from Lynx by loading its pretrained weights. Therefore, in Table \ref{tab:res_RGB}, we compare AllSpark with the state-of-the-art models. The results indicate that AllSpark outperforms most baseline models, with a margin of only 0.84 compared to that of the SOTA (95.69). This highlights AllSpark's exceptional perception and interpretation capabilities in the RGB modality.

\begin{table}[htbp]
  \centering
  \caption{RGB image classification with allspark}
    \begin{tabular}{p{0.48\columnwidth}p{0.18\columnwidth}c}
    \toprule
    Method & Publication & Acc(\%) \\
    \midrule
    CNN-CapsNet\cite{zhang2019remote} & RS2019 & 89.03 \\
    DFAGCN\cite{xu2021deep} & TNNLS2021 & 89.29 \\
    D-CNN with GoogleNet\cite{cheng2018deep} & TGRS2018 & 90.49 \\
    D-CNN with VGGNet\cite{cheng2018deep} & TGRS2018 & 91.89 \\
    SCCov\cite{he2019skip} & TNNLS2019 & 92.1 \\
    SeCo-ResNet-50\cite{manas2021seasonal} & ICCV2021 & 92.91 \\
    MG-CAP\cite{wang2020multi} & TIP2020 & 92.95 \\
    LSENet\cite{bi2021local} & TIP2021 & 93.34 \\
    MSANet\cite{zhang2021multiscale} & JSTARS2021 & 93.52 \\
    IDCCP\cite{wang2020invariant} & TGRS2021 & 93.76 \\
    MBLANet\cite{chen2021remote} & TIP2021 & 94.66 \\
    GRMANet-ResNet-50\cite{li2021gated} & TGRS2021 & 94.72 \\
    EMSNet\cite{zhao2023emscnet} & TGRS2023 & 95.37 \\
    ViTAE-B + RVSA\cite{wang2022advancing} & TGRS2022 & \textbf{95.69} \\
    \midrule
    AllSpark & ours  & 94.85 \\
    \bottomrule
    \end{tabular}%
  \label{tab:res_RGB}%
\end{table}%

\subsubsection{MSI}
\label{section4.3.2}

We evaluated the performance of AllSpark on the MSI scene classification task using the EuroSAT dataset. In the experiment, all 13 spectral bands of the images were simultaneously input into the model. The model's objective was to correctly classify the images into one of the 10 specified categories, and the evaluation metric chosen was the top-1 accuracy. AllSpark does not possess expert knowledge in the multispectral modality, so we categorize the baseline models into two groups, as shown in Table \ref{tab:res_MSI}: those with expert knowledge intervention and those without. Expert knowledge intervention refers to baseline models pretraining on large datasets such as BigEarthNet and then fine-tuning on the EuroSAT dataset, while no expert knowledge indicates baseline models trained directly from scratch on the EuroSAT dataset. The results show that our model outperforms most models in the no expert knowledge group, with a margin of only 2.60 compared to the state-of-the-art model (ResNet-152). Furthermore, AllSpark lags behind the best result in the expert knowledge group by only 4.75, demonstrating its excellent adaptability to the multispectral modality.

\begin{table}[htbp]
  \centering
  \caption{MSI land cover classification with allspark}
    \begin{tabular}{p{0.18\columnwidth}p{0.26\columnwidth}p{0.18\columnwidth}c}
    \toprule
    \multicolumn{1}{p{0.18\columnwidth}}{Expert Know.} & Method & Publication & \multicolumn{1}{c}{Acc(\%)} \\
    \midrule
    \multicolumn{1}{c}{\multirow{6}[2]{*}{\ding{55}}} & ResNet-18\cite{manas2021seasonal} & ICCV2021 & 63.21 \\
          & ResNet-50\cite{nampally2023performance} & IGARSS2023 & 91.13 \\
          & InceptionNet\cite{zhang2020semi} & ICIP2020 & 93.07 \\
          & EfficientNet\cite{gomez2021msmatch} & IJSTARS2021 & 93.94 \\
          & \underline{\textbf{AllSpark}} & \underline{\textbf{ours}} & \underline{94.03} \\
          & ResNet-152\cite{nampally2023performance} & IGARSS2023 & \textbf{96.63} \\
    \midrule
    \multicolumn{1}{c}{\multirow{4}[2]{*}{\ding{52}}} & MoCoV2\cite{manas2021seasonal} & ICCV2021 & 89.51 \\
          & SeCo\cite{manas2021seasonal} & ICCV2021 & 93.14 \\
          & SEER\cite{goyal2022vision} & Arxiv2022 & 97.6 \\
          & DINO-MC\cite{wanyan2023dino} & Arxiv2023 & \textbf{98.78} \\
    \bottomrule
    \end{tabular}%
  \label{tab:res_MSI}%
\end{table}%

\subsubsection{HSI}
\label{section4.3.3}

We conducted a pixel classification task on the Pavia University dataset for the hyperspectral modality. The model treats all spectral bands of a single pixel as one sample and predicts the land cover category of that pixel. The reported metrics include overall accuracy (OA), average accuracy (AA), and kappa. Since AllSpark does not possess expert knowledge of the hyperspectral modality, we compared it with the semisupervised baselines summarized by D. Uchaev and D. Uchaev \cite{uchaev2023small}. The results in Table \ref{tab:res_HSI} demonstrate that AllSpark outperforms many hyperspectral image classification methods, such as IFRF and S-DMM, by a factor of 6.42 compared to the best result in terms of OA, highlighting AllSpark's superior adaptability to the hyperspectral modality.

\begin{table}[htbp]
  \centering
  \caption{HSI pixel classification with allspark}
    \begin{tabular}{lccc}
    \toprule
    \multicolumn{1}{l}{Method} & \multicolumn{1}{p{0.15\columnwidth}<{\centering}}{OA} & \multicolumn{1}{p{0.15\columnwidth}<{\centering}}{AA} & \multicolumn{1}{p{0.15\columnwidth}<{\centering}}{Kappa} \\
    \midrule
    3D-CNN & 75.24 & 80.26 & 68.34 \\
    CA-GAN & 76.81 & 76.94 & 71.02 \\
    3D VS-CNN & 81.63 & 83.86 & 76.46 \\
    RPNet & 84.92 & 83.26 & 80.52 \\
    S-DMM & 88.3  & 93.76 & 84.9 \\
    IFRF  & 88.38 & 85.99 & 84.97 \\
    \underline{\textbf{AllSpark}} & \underline{89.18} & \underline{86.65} & \underline{85.32} \\
    DCFSL & 90.71 & 90.2  & 87.73 \\
    TC-GAN & 93.2  & 91.6  & 91 \\
    PRNet-RF & \textbf{95.6} & \textbf{94.96} & \textbf{94.27} \\
    \bottomrule
    \end{tabular}%
  \label{tab:res_HSI}%
\end{table}%

\subsubsection{Table}
\label{section4.3.4}

For the table modality, we evaluated AllSpark on the regression prediction task using the PRSA\cite{zhang2017cautionary} dataset. The task involves predicting the concentration of PM2.5 in the air using features such as time, site, 4 air pollutants, and 6 meteorological variables (as detailed in Section \ref{section4.2}). The performance metrics include the root mean squared error (RMSE), mean absolute error (MAE), and R-squared (R2).

\begin{table}[htbp]
  \centering
  \caption{PM2.5 prediction with allspark}
    \begin{tabular}{clccc}
    \toprule
    \multicolumn{1}{p{0.15\columnwidth}}{\multirow{2}[4]{*}{Expert Arch.}} & \multirow{2}[4]{*}{Method} & \multicolumn{3}{c}{PM2.5} \\
\cmidrule{3-5}          & \multicolumn{1}{r}{} & \multicolumn{1}{c}{RMSE} & \multicolumn{1}{c}{MAE} & \multicolumn{1}{c}{R2} \\
    \midrule
    \multicolumn{1}{c}{\multirow{3}[2]{*}{\ding{55}}} & GWO\cite{xie2020evolving} & 62.2  & 40.8  & \multicolumn{1}{c}{-} \\
          & TabBERT\cite{padhi2021tabular} & 32.8  & \multicolumn{1}{c}{-} & \multicolumn{1}{c}{-} \\
          & \textbf{AllSpark} & \textbf{29.03} & \textbf{18.04} & \textbf{0.87} \\
    \midrule
    \multicolumn{1}{c}{\multirow{2}[2]{*}{\ding{52}}} & Stacked ResNet-LSTM\cite{cheng2022stacked} & 40.68 & 23.75 & 0.8 \\
          & CBAM-CNN-Bi-LSTM\cite{li2022prediction} & \textbf{18.9} & \textbf{11.2} & \textbf{0.94} \\
    \bottomrule
    \end{tabular}%
  \label{tab:res_table}%
\end{table}%

Table \ref{tab:res_table} presents the comparative results between AllSpark and the baselines. It is worth noting that some works specifically focus on the prediction task on the PRSA dataset and design expert models with specific architectures based on dataset characteristics. For example, the CBAM-CNN-Bi-LSTM proposed by D. Li et al.\cite{li2022prediction} used a CNN to extract spatial dependencies between air monitoring stations and Bi-LSTM to capture the temporal dependencies of PM2.5 data. Similarly, the stacked ResNet-LSTM model proposed by X. Cheng et al.\cite{cheng2022stacked} employs a stacking LSTM strategy to enhance the extraction of temporal features in PM2.5 data. Therefore, we categorize baseline methods into two types: those with an expert architecture and those without an expert architecture. Among the models without expert architecture, our approach achieves the best performance among the baselines and is slightly inferior to the state-of-the-art method for models with expert architecture (CBAM-CNN-Bi-LSTM). This reflects the excellent adaptability of AllSpark to the table modality.

\subsubsection{Code}
\label{section4.3.5}

For the code modality, we evaluated the performance of AllSpark on the code document generation task using the CodeSearchNet dataset. This task involves generating corresponding documents based on the provided function code. We conducted tests for both the Ruby and JavaScript languages using the mean reciprocal rank (MRR)\cite{guo2020graphcodebert} as the evaluation metric.

\begin{table}[htbp]
  \centering
  \caption{code document generation with allspark}
    \begin{tabular}{p{0.19\columnwidth}lcc}
    \toprule
    \multicolumn{1}{c}{\multirow{2}[4]{*}{Expert Know.}} & \multirow{2}[4]{*}{Method} & \multicolumn{2}{c}{MRR} \\
\cmidrule{3-4}          & \multicolumn{1}{r}{} & \multicolumn{1}{c}{Ruby} & \multicolumn{1}{c}{Javascript} \\
    \midrule
    \multicolumn{1}{c}{\multirow{6}[2]{*}{\ding{55}}} & BIRNN[68] & 0.084 & 0.153 \\
          & 1D-CNN[68] & 0.245 & 0.352 \\
          & selfAtt[68] & 0.365 & 0.451 \\
          & NBoW[68] & 0.429 & 0.461 \\
          & RoBERTa[67] & 0.625 & 0.606 \\
          & \textbf{AllSpark} & \textbf{0.627} & \textbf{0.635} \\
    \midrule
    \multicolumn{1}{c}{\multirow{3}[2]{*}{\ding{52}}} & RoBERTa(Code)[67] & 0.661 & 0.64 \\
          & CodeBERT[67] & 0.693 & 0.706 \\
          & GraphCodeBERT[67] & \textbf{0.732} & \textbf{0.711} \\
    \bottomrule
    \end{tabular}%
  \label{tab:res_code}%
\end{table}%

Like in the previous modalities, since MSI-AGI does not possess expert knowledge of the code modality, we categorized the baselines into two groups: those with and without expert knowledge. As shown in Table \ref{tab:res_code}, AllSpark achieved SOTA results in the group without expert knowledge, and the results were comparable to those of models trained with expert knowledge. This finding demonstrates the strong adaptability of AllSpark to the code modality.

\subsubsection{Point Cloud}
\label{section4.3.6}

For the point cloud modality, we evaluated the performance of AllSpark on the ModelNet40 dataset for the classification task, with the top-1 accuracy as the metric. In the context of single-modal studies focused on point clouds, we observed that due to the unique three-dimensional structure of point cloud data, most works concentrate on designing specific structures to maintain properties such as permutation invariance and symmetry in 3D point clouds. However, these structures designed for the unique priors of the modality are challenging to transfer across modalities. Additionally, some methods tend to pretrain on large point cloud datasets to acquire general modal expert knowledge before generalizing to specific downstream tasks to improve performance.

\begin{table}[htbp]
  \centering
  \caption{pointcloud classification with allspark}
    \begin{tabular}{p{0.25\columnwidth}ccc}
    \toprule
    Method & \makecell[c]{Expert\\Architecture} & \makecell[c]{Expert\\Knowledge} & \multicolumn{1}{c}{Acc(\%)} \\
    \midrule
    PointNet\cite{qi2017pointnet} & \ding{52}   & \ding{55}    & 89.2 \\
    Kd-net\cite{klokov2017escape} & \ding{52}   & \ding{55}    & 90.6 \\
    SPH3D-GCN\cite{lei2020spherical} & \ding{52}   & \ding{55}    & 91.4 \\
    PointNet++\cite{qi2017pointnet++} & \ding{52}   & \ding{55}    & 91.9 \\
    SO-Net\cite{li2018so} & \ding{52}   & \ding{55}    & 92.5 \\
    PointVGG\cite{li2021pointvgg} & \ding{52}   & \ding{55}    & 93.6 \\
    PointBERT\cite{yu2022point} & \ding{55} & \ding{52}   & 93.8 \\
    PointGPT\cite{chen2023pointgpt} & \ding{52}   & \ding{52}   & \textbf{94.9} \\
    \midrule
    AllSpark & \ding{55} & \ding{55} & 91.2 \\
    \bottomrule
    \end{tabular}%
  \label{tab:res_pointcloud}%
\end{table}%

The AllSpark module is designed based on a general sequence-to-sequence architecture. As shown in Table \ref{tab:res_pointcloud}, in the absence of both modality expert architectural designs and modal expert knowledge, AllSpark still outperforms classical networks with point cloud-specific structures (PointNet, Kd-Net). This approach maintains comparability with the state-of-the-art PointGPT model, which has both a modal expert structure and modal expert knowledge. This finding suggested that AllSpark has significant potential for applications in the point cloud modality.

\subsubsection{Trajectory}
\label{section4.3.7}

For the trajectory modality, we evaluated the performance of AllSpark on the ETH dataset in the trajectory prediction task. This task involves predicting possible two-dimensional trajectories based on a set of two-dimensional coordinate points within a certain time period. We report accuracy using the average displacement error (ADE) and final displacement error (FDE)\cite{shi2023trajectory}. Given the future trajectory$\{x_t,y_t\}_{t=T_{bos}+1}^T$ (ground truth) and the predicted trajectory $\{\widehat{x_t},\widehat{y_t}\}_{t=T_{bos}+1}^T$, the ADE and FDE are used to measure their $L2$ distances, calculated as follows:

\begin{equation}
{ADE}=\frac{1}{T_{pred}}\sum_{t=T_{bos}}^{T}\sqrt{\left(x_t-\widehat{x_t}\right)^2+\left(y_t-\widehat{y_t}\right)^2}
\end{equation}

\begin{equation}
{FDE}=\sqrt{\left(x_T-\widehat{x_T}\right)^2+\left(y_T-\widehat{y_T}\right)^2}
\end{equation}

In Table \ref{tab:res_traj}, AllSpark is compared with the state-of-the-art trajectory prediction models. The results indicate that AllSpark, which uses a unified structure without trajectory modality expert knowledge, outperforms most expert models. It achieves a prediction accuracy close to that of the SOTA model (STAR), with a difference of only 0.07 in the ADE metric and 0.11 in the FDE metric. This finding suggested that AllSpark demonstrated excellent adaptability to the trajectory modality.

\begin{table}[htbp]
  \centering
  \caption{trajectory prediction with allspark}
    \begin{tabular}{p{8.5em}c}
    \toprule
    Method & ADE/FDE \\
    \midrule
    Social GAN & 0.87/1.62 \\
    SoPhie & 0.70/1.43 \\
    STAR  & \textbf{0.36}/0.64 \\
    SGCN  & 0.63/1.03 \\
    CAGN  & 0.41/0.65 \\
    SIT   & 0.39/0.62 \\
    SocialVAE & 0.47/0.76 \\
    PCENet & 0.54/0.87 \\
    AgentFormer & 0.45/0.75 \\
    MemoNet & 0.40/0.61 \\
    SocialVAE+FPC & 0.41/\textbf{0.58} \\
    TUTR  & 0.40/0.61 \\
    \midrule
    AllSpark & 0.43/0.69 \\
    \bottomrule
    \end{tabular}%
  \label{tab:res_traj}%
\end{table}%

\subsubsection{SAR}
\label{section4.3.8}

For the SAR modality, the adaptability of AllSpark was tested on the MSTAR dataset, where the model is required to identify SAR images of ten military targets, and the metric used is the top-1 accuracy. In the experiment, the preprocessing of the MSTAR dataset followed the SOC settings from AConvNets\cite{chen2016target}. Table \ref{tab:res_SAR} presents the comparison results between AllSpark and the state-of-the-art model under these settings. AllSpark achieves 97.24\% top1-accuracy, only 1.89\% lower than the SOTA model. The experiments demonstrate that AllSpark is capable of effectively understanding SAR imagery.

\begin{table}[htbp]
  \centering
  \caption{sar classification with allspark}
    \begin{tabular}{p{9.815em}c}
    \toprule
    Method & \multicolumn{1}{p{6.25em}<{\centering}}{Acc(\%)} \\
    \midrule
    EMACH\cite{srinivas2014sar} & 88 \\
    SVM\cite{srinivas2014sar} & 90 \\
    AdaBoost\cite{srinivas2014sar} & 92 \\
    MSRC\cite{dong2014sparse} & 93.6 \\
    IGT\cite{srinivas2014sar} & 95 \\
    MSS\cite{dong2015classification} & 96.6 \\
    Cond Gauss\cite{o2001sar} & 97 \\
    M-PMC\cite{park2014modified} & 98.8 \\
    AConvNets\cite{chen2016target} & \textbf{99.13} \\
    \midrule
    AllSpark & 97.24 \\
    \bottomrule
    \end{tabular}%
  \label{tab:res_SAR}%
\end{table}%

\subsubsection{Graph}
\label{section4.3.9}

For the graph modality, the performance of AllSpark was evaluated on the traffic flow prediction task using the METR-LA dataset. The evaluation metrics include the RMSE, MAE, and R2. Table \ref{tab:res_graph} compares AllSpark and the state-of-the-art methods on the METR-LA dataset, with the baseline derived from\cite{liu2023spatio}. The results show that AllSpark, without the intervention of modal expert knowledge, is only 0.47 away from the best result in terms of the RMSE, demonstrating its excellent adaptability to the graph modality.

\subsubsection{Language}
\label{section4.3.10}

We tested AllSpark's natural language processing capabilities on the IMDB dataset\cite{maas2011learning} with the task of binary sentiment classification (positive or negative). AllSpark loads weights from the Lynx; therefore, it can be considered to possess expert knowledge in the natural language modality. Compared with the SOTA models on the IMDB dataset, as shown in Table \ref{tab:res_language}, AllSpark outperformed most language models, trailing the SOTA result by only 0.32. This highlights AllSpark's powerful understanding and analysis capabilities in natural language.

\begin{table}[htbp]
  \centering
  \caption{traffic prediction with allspark}
    \begin{tabular}{p{5.875em}ccc}
    \toprule
    Method & \multicolumn{1}{c}{RMSE} & \multicolumn{1}{c}{MAE} & \multicolumn{1}{c}{R2} \\
    \midrule
    HI    & 6.8   & 14.2  & 10.15 \\
    GWNet & 3.51  & 7.28  & 9.96 \\
    DCRNN & 3.54  & 7.47  & 10.32 \\
    AGCRN & 3.59  & 7.45  & 10.47 \\
    STGCN & 3.6   & 7.43  & 10.35 \\
    GTS   & 3.59  & 7.44  & 10.25 \\
    MTGNN & 3.47  & 7.21  & 9.7 \\
    STNorm & 3.57  & 7.51  & 10.24 \\
    GMAN  & 3.44  & 7.35  & 10.07 \\
    PDFormer & 3.62  & 7.47  & 10.91 \\
    STID  & 3.55  & 7.55  & 10.95 \\
    STAEformer & \textbf{3.34} & \textbf{7.02} & \textbf{9.7} \\
    \midrule
    AllSpark & 3.81  & 7.52  & 11.24 \\
    \bottomrule
    \end{tabular}%
  \label{tab:res_graph}%
\end{table}%

\begin{table}[htbp]
  \centering
  \caption{text understanding with allspark}
    \begin{tabular}{p{9.625em}c}
    \toprule
    Method & \multicolumn{1}{p{6.625em}<{\centering}}{Acc(\%)} \\
    \midrule
    RoBERTa\cite{liu2019roberta} & 95.3 \\
    ULMFiT\cite{howard2018universal} & 95.4 \\
    BERT\cite{xie2020unsupervised} & 95.49 \\
    Mixed VAT\cite{sachan2019revisiting,miyato2016adversarial} & 95.68 \\
    LongFormer\cite{beltagy2020longformer} & 95.7 \\
    XLNet\cite{yang2019xlnet} & 96.8 \\
    ERNIE-Doc-Large\cite{ding2020ernie} & \textbf{97.1} \\
    \midrule
    AllSpark & 96.78 \\
    \bottomrule
    \end{tabular}%
  \label{tab:res_language}%
\end{table}%

\subsection{Training and Inference Costs}
\label{section4.5}

As the scale of model parameters grows, the training and inference costs of large language models have been rapidly increasing. In this section, we provide detailed the training and inference costs of AllSpark for reference. All experiments are conducted on two NVIDIA A6000 48G GPUs, using a numerical precision of torch.float32. The hyperparameter settings are detailed in Section \ref{section4.1}.

Due to the presence of modality-specific encoders, the activated parameters in AllSpark varies when processing different modalities. Table \ref{tab:cost} summarizes the total parameters, trainable parameters, training time, and inference computational cost (measured in MACs) for each modality.

\begin{table*}[]
\centering
\caption{training and inference costs}
\label{tab:cost}
\resizebox{0.85\textwidth}{!}{%
\begin{tabular}{ccccc}
\hline
Modality   & Total params.(M) & Trainable params.(M) & Training time  & Inference MACs(G) \\ \hline
RGB        & 8176.02          & 689.80               & $\sim$18 hours & 1695.98           \\
MSI        & 7455.14          & 978.87               & $\sim$30 hours & 363.02            \\
HSI        & 7450.58          & 974.31               & $\sim$21 hours & 374.61            \\
Table      & 7486.67          & 1010.40              & $\sim$44 hours & 410.00            \\
Code       & 6876.02          & 268.66               & $\sim$59 hours & 3606.88           \\
Point      & 7594.81          & 1118.54              & $\sim$49 hours & 557.38            \\
Trajectory & 7381.00          & 904.73               & $\sim$39 hours & 538.58            \\
SAR        & 7452.58          & 976.31               & $\sim$87 hours & 373.63            \\
Graph      & 7223.91          & 747.64               & $\sim$53 hours & 1846.28           \\
Language   & 6876.02          & 268.66               & $\sim$34 hours & 3606.88           \\ \hline
\end{tabular}%
}
\end{table*}

\section{Discussion}
\label{section5}

\subsection{Limitations}
\label{section5.1}

Certainly, our work has several limitations, which will guide our future research directions:

\begin{enumerate}
    \item \textbf{Lack of interaction between different modalities}. AllSpark only facilitates interactions between the language and other modalities, without involving more interactions, such as RGB and point cloud, or hyperspectral and multispectral. This is primarily due to constraints such as the lack of multimodal paired data. However, collecting large-scale paired data for ten modalities is nearly impossible, so we attempt to use language as the alignment reference for each modality, achieving indirect alignment between modalities using unpaired data. AllSpark represents the initial effort in this approach and has demonstrated strong adaptability across various modalities. Also, the adversarial examples also affect the robustness of the proposed model \cite{Cui2024AEPT}. In the future, we plan to explore and expand our efforts in these directions in the further.
    \item \textbf{Initial work}. Our current work is still in its initial exploratory phase, and we have not carefully refined AllSpark's adaptability and performance on each modality. As a result, the model exhibits suboptimal performance on certain modalities, such as oblique photography, and video. The experimental results can be found in Table \ref{tab:res_oblique} and Table \ref{tab:res_video}. In the future, we plan to conduct more refined and targeted adjustments for each modality to enhance overall performance.
    \item \textbf{Expensive cost}: Multimodal large language models, due to prolonged pretraining, typically possess universal reasoning capabilities in certain modalities. We generalize their applicability to other modalities by utilizing modality bridges to project other modalities onto the language modality. As shown in Section \ref{section4.5}, although we freeze most of the parameters, fine-tuning even once on 2 A6000 GPUs often requires more than a day. Given the increasing training costs for large models, exploring methods to generalize their universal reasoning abilities is one of our future research directions.
    \item \textbf{Interesting phenomenon}. During our exploratory experiments, we discovered several interesting properties of large models. For instance, spatial information tends to degrade in large models, leading to collapse when performing dense prediction tasks such as segmentation and detection. Additionally, these models struggle to optimize when transferred to a small quantity of downstream data. Large models often require smaller hyperparameters and are sensitive to them. These observations might partially reveal the working mechanisms of large models, and we plan to conduct additional in-depth investigations into these phenomena in the future.
\end{enumerate}

\subsection{Potential of the LaRF}
\label{section5.2}

Inspired by the human cognitive system and linguistic philosophy, we propose the "LaRF" as the first principle for constructing our unified multimodal model. Its foreseeable potential includes at least the following three points:

\begin{enumerate}
    \item \textbf{Efficient Generalization of Large Models}. Currently, the computational power and data scale required for training large models are rapidly expanding, and even the cost of fine-tuning large models is becoming prohibitive. Therefore, in the future, training large models for every domain will be almost impossible. Language, however, holds the potential for achieving efficient generalization of large models, expanding them from their native domains to additional domains at minimal cost. With our proposed AllSpark, we designed simple text prompts and task heads for each modality, demonstrating significant potential for multimodal expansion. In theory, AllSpark, built on the LaRF principle, can be extended to arbitrary modalities. In the future, we will conduct more in-depth research on the impact of text prompts and lightweight parameter modules on the generalization of large models.
    \item \textbf{Interpretable Reasoning}. Deep learning models have often been referred to as "black boxes," indicating that the reasoning process of these models is invisible and challenging to interpret. Research on the interpretability of deep learning models often relies on complex mathematical models and numerous assumptions, greatly limiting the practical application of deep learning methods in fields such as clinical medicine, military, and national resources where low fault tolerance or high confidentiality are crucial. However, language, as a tool for human thought and communication, provides models based on LaRF with the potential to use natural language directly for outputting reasoning chains and justifications.
    \item \textbf{Transition from an end-to-end to an interactive paradigm}. The end-to-end paradigm refers to the learning approach where the model takes input and directly outputs results. In recent years, the end-to-end paradigm has become increasingly popular due to its simple and clear architecture and excellent performance. However, this approach also has clear disadvantages, such as uncontrollable internal operations, the need to optimize the whole for certain problems, and difficulty in pinpointing the cause of issues. A LaRF-based architecture has the potential to achieve an interactive paradigm in which users input raw data and corresponding text prompts and the model automatically performs relevant operations based on the prompts. Users can even iteratively adjust the text prompts based on the results. Therefore, in terms of both performance and controllability, the interactive paradigm has advantages that are incomparable to those of the end-to-end paradigm.
\end{enumerate}

\section{Conclusion}
\label{section6}

Leveraging multimodal data is an inherent requirement for intelligent models to achieve geographic object cognition. Inspired by human cognitive systems and linguistic philosophy, we propose that the construction of multimodal models follow the fundamental principle of 
\textbf{L}anguage \textbf{a}s \textbf{R}eference \textbf{F}ramework (LaRF). Guided by this principle, we use language to balance the cohesion and autonomy of modalities, presenting a unified intelligent model, AllSpark, encompassing ten spatio-temporal modalities. The experimental results demonstrated that AllSpark exhibited excellent adaptability and application potential across various spatio-temporal modalities, highlighting the feasibility and potential of constructing multimodal models with LaRF. AllSpark remains an initial exploratory work, and in the future, we aim to delve deeper into the mechanisms guided by natural language, the efficient generalization of large models, and the transition to an interactive paradigm.

\appendix
\section{Appendix}

\subsection{List of prompts}

The text prompts we used are listed in Table \ref{tab:prompts}.

\begin{table*}[!t]
\centering 
\caption{List of prompts} 
\label{tab:prompts} 
\resizebox{0.85\textwidth}{!}{%
\begin{tabular}{ccl}
\hline
Dimension &
  Modal &
  \multicolumn{1}{c}{Prompts} \\ \hline
\multirow{3}[8]{*}{1D} &
  Language &
  1.Please determine if this movie review is positive or negative? \\ \cline{2-3} 
 &
  Code &
  - \\ \cline{2-3} 
 &
  Table &
  \begin{tabular}[c]{@{}l@{}}1.Please utilize the provided air quality indicators to accurately\\ predict the concentrations of PM2.5 in the atmosphere\\ 2.Given the air quality indicators, please provide a prediction\\ for the PM2.5 levels at this particular moment\end{tabular} \\ \hline
\multirow{8}[55]{*}{2D} &
  RGB &
  \begin{tabular}[c]{@{}l@{}}1.This remote sensing image belongs to which of the following\\ categories: {[}Category list of the NWPU dataset{]}\\ 2.Describe this remote sensing image briefly3.Find a word that\\ is most relevant to this remote sensing image\\ 3.Find a word that is most relevant to this remote sensing image\\ 4.Describe the key elements in this remote sensing image\end{tabular} \\ \cline{2-3} 
 &
  MSI &
  \begin{tabular}[c]{@{}l@{}}1.Based on the multi-spectral imagery feature description, please\\ classify this object\\ 2.Given the following multi-spectral imagery characteristics, please\\ output the most fitting scene label\end{tabular} \\ \cline{2-3} 
 &
  HSI &
  \begin{tabular}[c]{@{}l@{}}1.Given the spectral information, can you help determine which\\ class this pixel belongs to?\\ 2.Here is the spectral data for a pixel. Considering the typical\\ characteristics of land cover classes, could you provide a detailed\\ analysis and suggest the most likely class for this pixel?\\ 3.The spectral information for a pixel is given, but the data is\\ noisy. Given the potential variability, which land cover classes\\ should be considered as possible candidates for this pixel?\end{tabular} \\ \cline{2-3} 
 &
  Trajectory &
  \begin{tabular}[c]{@{}l@{}}1.Based on their past positions and movements in a crowded\\ environment, predict the future trajectory of a selected pedestrian\\ 2.Using the pedestrian trajectory data, along with additional\\ information about the surrounding environment, predict the\\ future path of the pedestrian\\ 3.Given the current and past positions of a pedestrian and their\\ neighboring pedestrians, predict the main pedestrian's trajectory\end{tabular} \\ \cline{2-3} 
 &
  SAR &
  \begin{tabular}[c]{@{}l@{}}1.Based on the SAR imagery feature description, please classify\\ this object\\ 2.Given the following SAR imagery characteristics, please output\\ the most fitting scene label\end{tabular} \\ \cline{2-3} 
 &
  Graph &
  \begin{tabular}[c]{@{}l@{}}1.Given the current traffic data including vehicle flow rate, average\\ speed, and time of day from the METRLA dataset, predict the\\ traffic flow\\ 2.Analyze the historical data on vehicle speeds and flow rates from\\ the METRLA dataset for the past week. Identify any patterns or\\ trends and predict the traffic conditions\end{tabular} \\ \hline
3D &
  Point Cloud &
  \begin{tabular}[c]{@{}l@{}}1.Classify the provided point cloud sample into the correct category\\ 2.Look at the point cloud data characteristics and classify the object\\ 3.Please analyze the given point cloud dataset and determine which\\ category it belongs to. Focus on the shape and structure evident in\\ the point cloud\end{tabular} \\ \hline
\end{tabular}%
}
\end{table*}

\subsection{Video}
\label{section:video}

For the video modality, we evaluated AllSpark's performance on action recognition tasks using the UCF101\cite{soomro2012ucf101} dataset. The UCF101 dataset is a human action recognition dataset comprising 101 action classes with a total of 13,320 video clips. The videos have a combined duration of 27 hours and a resolution of 320*240 pixels and were sourced from YouTube.

The model is tasked with understanding videos and accurately classifying them into one of the 101 classes, with the evaluation metric being the top-1 accuracy. In Table \ref{tab:res_video}, we compare AllSpark with the current state-of-the-art models. 

Currently, AllSpark's adaptation to the video modality is not optimal, as it shows a significant difference from the baseline model results. We attribute this to two main reasons: 1. The high redundancy in video information increases the training cost for AllSpark. We trained it for only 3 epochs on the dataset. 2. AllSpark's model architecture lacks flexibility for three-dimensional data, making it less effective at capturing temporal information.

\begin{table}[htbp]
  \centering
  \caption{video classification with allspark}
    \begin{tabular}{p{6.875em}c}
    \toprule
    Method & \multicolumn{1}{c}{Acc(\%)} \\
    \midrule
    OPN   & 59.6 \\
    VCOP  & 72.4 \\
    SpeedNet & 81.1 \\
    VTHCL & 82.1 \\
    CVRL  & 94.4 \\
    VideoMAE v1 & 96.1 \\
    VideoMAE v2 & 99.6 \\
    \midrule
    AllSpark & 27.5 \\
    \bottomrule
    \end{tabular}%
  \label{tab:res_video}%
\end{table}%

\subsection{Oblique Photography}
\label{section:OP}

For the oblique photography modality, we tested AllSpark's performance on the 3D reconstruction task using the WHU-OMVS\cite{liu2023deep} dataset. WHU-OMVS is an oblique photography dataset designed for 3D reconstruction tasks. The dataset provides imagery from five different viewpoints, along with camera parameters and other relevant information. It consists of six areas, and in our experiments, area 1 is used as the training set, and area 2 is used as the test set.

The model takes five-view images as input, and the goal is to output depth maps for reconstructing 3D models. The evaluation metric used was the percentage of accurate grids in total (PAG)\cite{liu2023deep}, calculated by the following formula:

\begin{equation}
PAG_a=\left(\frac{m_a}{m}\right)*100%
\end{equation}

The suffixes in the PAG represent different accuracy standards, where ${PAG}_6$ signifies an error within 0.6 metres and ${PAG}_{10}$ indicates an error within 1 metre.
Table \ref{tab:res_oblique} compares AllSpark with popular multiview 3D reconstruction models on the WHU-OMVS dataset. 

AllSpark falls short in terms of accuracy compared to modality-specific expert models. We speculate two possible reasons: 1. Dense spatial information gradually diminishes in the deep structure of large models, a point verified in our exploratory experiments involving segmentation, detection, etc.; 2. The model architecture lacks flexibility and cannot connect gradual features like expert 3D reconstruction models such as Ada-MVS. Additionally, these methods lack the ability to design specific model structures for processing, severely restricting their performance. Adapting multimodal large models to dense prediction tasks and optimizing the architecture are future research directions.

\begin{table}[htbp]
  \centering
  \caption{3D reconstruction with allspark}
    \begin{tabular}{p{6.25em}cc}
    \toprule
    Method & \multicolumn{1}{p{4.19em}<{\centering}}{${PAG}_6$} & \multicolumn{1}{p{4.19em}<{\centering}}{${PAG}_{10}$} \\
    \midrule
    MVSNet & 81.15 & 91.44 \\
    CasMVSNet & 95.45 & 98.02 \\
    Ada-MVS & 96.14 & 98.1 \\
    UCSNet & 96.25 & 98.45 \\
    \midrule
    AllSpark & 6.4   & 10.4 \\
    \bottomrule
    \end{tabular}%
  \label{tab:res_oblique}%
\end{table}%

\bibliographystyle{IEEEtran}
\bibliography{ref}

\begin{thebibliography}{100}
\providecommand{\url}[1]{#1}
\csname url@samestyle\endcsname
\providecommand{\newblock}{\relax}
\providecommand{\bibinfo}[2]{#2}
\providecommand{\BIBentrySTDinterwordspacing}{\spaceskip=0pt\relax}
\providecommand{\BIBentryALTinterwordstretchfactor}{4}
\providecommand{\BIBentryALTinterwordspacing}{\spaceskip=\fontdimen2\font plus
\BIBentryALTinterwordstretchfactor\fontdimen3\font minus \fontdimen4\font\relax}
\providecommand{\BIBforeignlanguage}[2]{{%
\expandafter\ifx\csname l@#1\endcsname\relax
\typeout{** WARNING: IEEEtran.bst: No hyphenation pattern has been}%
\typeout{** loaded for the language `#1'. Using the pattern for}%
\typeout{** the default language instead.}%
\else
\language=\csname l@#1\endcsname
\fi
#2}}
\providecommand{\BIBdecl}{\relax}
\BIBdecl

\bibitem{zhang2024earthgpt}
W.~Zhang, M.~Cai, T.~Zhang, Y.~Zhuang, and X.~Mao, ``Earthgpt: A universal multi-modal large language model for multi-sensor image comprehension in remote sensing domain,'' \emph{IEEE Transactions on Geoscience and Remote Sensing}, 2024.

\bibitem{guo2024skysense}
X.~Guo, J.~Lao, B.~Dang, Y.~Zhang, L.~Yu, L.~Ru, L.~Zhong, Z.~Huang, K.~Wu, D.~Hu \emph{et~al.}, ``Skysense: A multi-modal remote sensing foundation model towards universal interpretation for earth observation imagery,'' in \emph{Proceedings of the IEEE/CVF Conference on Computer Vision and Pattern Recognition}, 2024, pp. 27\,672--27\,683.

\bibitem{xu2023pointllm}
R.~Xu, X.~Wang, T.~Wang, Y.~Chen, J.~Pang, and D.~Lin, ``Pointllm: Empowering large language models to understand point clouds,'' \emph{arXiv preprint arXiv:2308.16911}, 2023.

\bibitem{chen-etal-2023-tablevlm}
\BIBentryALTinterwordspacing
L.~Chen, C.~Huang, X.~Zheng, J.~Lin, and X.~Huang, ``{T}able{VLM}: Multi-modal pre-training for table structure recognition,'' in \emph{Proceedings of the 61st Annual Meeting of the Association for Computational Linguistics (Volume 1: Long Papers)}, A.~Rogers, J.~Boyd-Graber, and N.~Okazaki, Eds.\hskip 1em plus 0.5em minus 0.4em\relax Toronto, Canada: Association for Computational Linguistics, Jul. 2023, pp. 2437--2449. [Online]. Available: \url{https://aclanthology.org/2023.acl-long.137}
\BIBentrySTDinterwordspacing

\bibitem{han2023onellm}
J.~Han, K.~Gong, Y.~Zhang, J.~Wang, K.~Zhang, D.~Lin, Y.~Qiao, P.~Gao, and X.~Yue, ``Onellm: One framework to align all modalities with language,'' \emph{arXiv preprint arXiv:2312.03700}, 2023.

\bibitem{zhang2023meta}
Y.~Zhang, K.~Gong, K.~Zhang, H.~Li, Y.~Qiao, W.~Ouyang, and X.~Yue, ``Meta-transformer: A unified framework for multimodal learning,'' \emph{arXiv preprint arXiv:2307.10802}, 2023.

\bibitem{he2023STGC-GNNs}
S.~He, Q.~Luo, R.~Du, L.~Zhao, G.~He, H.~Fu, and H.~Li, ``Stgc-gnns: A gnn-based traffic prediction framework with a spatial–temporal granger causality graph,'' \emph{Physica A: Statistical Mechanics and its Applications}, vol. 623, p. 128913, 2023.

\bibitem{luo2024lsttn}
Q.~Luo, S.~He, X.~Han, Y.~Wang, and H.~Li, ``Lsttn: A long-short term transformer-based spatiotemporal neural network for traffic flow forecasting,'' \emph{Knowledge-Based Systems}, vol. 293, p. 111637, 2024.

\bibitem{li2024}
H.~Li, J.~Cao, J.~Zhu, Q.~Luo, S.~He, and X.~Wang, ``Augmentation-free graph contrastive learning of invariant-discriminative representations,'' \emph{IEEE Transactions on Neural Networks and Learning Systems}, vol.~35, no.~8, pp. 11\,157 -- 11\,167, 2024.

\bibitem{qi2017pointnet}
C.~R. Qi, H.~Su, K.~Mo, and L.~J. Guibas, ``Pointnet: Deep learning on point sets for 3d classification and segmentation,'' in \emph{Proceedings of the IEEE conference on computer vision and pattern recognition}, 2017, pp. 652--660.

\bibitem{vaswani2017attention}
A.~Vaswani, N.~Shazeer, N.~Parmar, J.~Uszkoreit, L.~Jones, A.~N. Gomez, {\L}.~Kaiser, and I.~Polosukhin, ``Attention is all you need,'' \emph{Advances in neural information processing systems}, vol.~30, 2017.

\bibitem{kipf2016semi}
T.~N. Kipf and M.~Welling, ``Semi-supervised classification with graph convolutional networks,'' \emph{arXiv preprint arXiv:1609.02907}, 2016.

\bibitem{li2021cgnn}
H.~Li, J.~Cao, J.~Zhu, Y.~Liu, Q.~Zhu, and G.~Wu, ``Curvature graph neural network,'' \emph{Information Sciences}, vol. 592, pp. 50--66, 2022.

\bibitem{kulkarni2020pixel}
S.~C. Kulkarni and P.~P. Rege, ``Pixel level fusion techniques for sar and optical images: A review,'' \emph{Information Fusion}, vol.~59, pp. 13--29, 2020.

\bibitem{radford2021learning}
A.~Radford, J.~W. Kim, C.~Hallacy, A.~Ramesh, G.~Goh, S.~Agarwal, G.~Sastry, A.~Askell, P.~Mishkin, J.~Clark \emph{et~al.}, ``Learning transferable visual models from natural language supervision,'' in \emph{International conference on machine learning}.\hskip 1em plus 0.5em minus 0.4em\relax PMLR, 2021, pp. 8748--8763.

\bibitem{lu2019vilbert}
J.~Lu, D.~Batra, D.~Parikh, and S.~Lee, ``Vilbert: Pretraining task-agnostic visiolinguistic representations for vision-and-language tasks,'' \emph{Advances in neural information processing systems}, vol.~32, 2019.

\bibitem{kim2021vilt}
W.~Kim, B.~Son, and I.~Kim, ``Vilt: Vision-and-language transformer without convolution or region supervision,'' in \emph{International Conference on Machine Learning}.\hskip 1em plus 0.5em minus 0.4em\relax PMLR, 2021, pp. 5583--5594.

\bibitem{alayrac2022flamingo}
J.-B. Alayrac, J.~Donahue, P.~Luc, A.~Miech, I.~Barr, Y.~Hasson, K.~Lenc, A.~Mensch, K.~Millican, M.~Reynolds \emph{et~al.}, ``Flamingo: a visual language model for few-shot learning,'' \emph{Advances in Neural Information Processing Systems}, vol.~35, pp. 23\,716--23\,736, 2022.

\bibitem{li2022blip}
J.~Li, D.~Li, C.~Xiong, and S.~Hoi, ``Blip: Bootstrapping language-image pre-training for unified vision-language understanding and generation,'' in \emph{International Conference on Machine Learning}.\hskip 1em plus 0.5em minus 0.4em\relax PMLR, 2022, pp. 12\,888--12\,900.

\bibitem{Tao2023TOV}
C.~Tao, J.~Qi, G.~Zhang, Q.~Zhu, W.~Lu, and H.~Li, ``Tov: The original vision model for optical remote sensing image understanding via self-supervised learning,'' \emph{IEEE Journal of Selected Topics in Applied Earth Observations and Remote Sensing}, vol.~16, pp. 4916--4930, 2023.

\bibitem{jaegle2021perceiver}
A.~Jaegle, F.~Gimeno, A.~Brock, O.~Vinyals, A.~Zisserman, and J.~Carreira, ``Perceiver: General perception with iterative attention,'' in \emph{International conference on machine learning}.\hskip 1em plus 0.5em minus 0.4em\relax PMLR, 2021, pp. 4651--4664.

\bibitem{shao2024}
R.~Shao, Z.~Zhang, C.~Tao, Y.~Zhang, C.~Peng, and H.~Li, ``Homogeneous tokenizer matters: Homogeneous visual tokenizer for remote sensing image understanding,'' \emph{ISPRS Journal of Photogrammetry and Remote Sensing}, vol.~18, pp. 294--310, 2024.

\bibitem{feng2020codebert}
Z.~Feng, D.~Guo, D.~Tang, N.~Duan, X.~Feng, M.~Gong, L.~Shou, B.~Qin, T.~Liu, D.~Jiang \emph{et~al.}, ``Codebert: A pre-trained model for programming and natural languages,'' \emph{arXiv preprint arXiv:2002.08155}, 2020.

\bibitem{kenton2019bert}
J.~D. M.-W.~C. Kenton and L.~K. Toutanova, ``Bert: Pre-training of deep bidirectional transformers for language understanding,'' in \emph{Proceedings of naacL-HLT}, vol.~1, 2019, p.~2.

\bibitem{radford2018improving}
A.~Radford, K.~Narasimhan, T.~Salimans, I.~Sutskever \emph{et~al.}, ``Improving language understanding by generative pre-training,'' 2018.

\bibitem{radford2019language}
A.~Radford, J.~Wu, R.~Child, D.~Luan, D.~Amodei, I.~Sutskever \emph{et~al.}, ``Language models are unsupervised multitask learners,'' \emph{OpenAI blog}, vol.~1, no.~8, p.~9, 2019.

\bibitem{brown2020language}
T.~Brown, B.~Mann, N.~Ryder, M.~Subbiah, J.~D. Kaplan, P.~Dhariwal, A.~Neelakantan, P.~Shyam, G.~Sastry, A.~Askell \emph{et~al.}, ``Language models are few-shot learners,'' \emph{Advances in neural information processing systems}, vol.~33, pp. 1877--1901, 2020.

\bibitem{achiam2023gpt}
J.~Achiam, S.~Adler, S.~Agarwal, L.~Ahmad, I.~Akkaya, F.~L. Aleman, D.~Almeida, J.~Altenschmidt, S.~Altman, S.~Anadkat \emph{et~al.}, ``Gpt-4 technical report,'' \emph{arXiv preprint arXiv:2303.08774}, 2023.

\bibitem{arik2021tabnet}
S.~{\"O}. Arik and T.~Pfister, ``Tabnet: Attentive interpretable tabular learning,'' in \emph{Proceedings of the AAAI conference on artificial intelligence}, vol.~35, no.~8, 2021, pp. 6679--6687.

\bibitem{he2016deep}
K.~He, X.~Zhang, S.~Ren, and J.~Sun, ``Deep residual learning for image recognition,'' in \emph{Proceedings of the IEEE conference on computer vision and pattern recognition}, 2016, pp. 770--778.

\bibitem{dosovitskiy2020image}
A.~Dosovitskiy, L.~Beyer, A.~Kolesnikov, D.~Weissenborn, X.~Zhai, T.~Unterthiner, M.~Dehghani, M.~Minderer, G.~Heigold, S.~Gelly \emph{et~al.}, ``An image is worth 16x16 words: Transformers for image recognition at scale,'' \emph{arXiv preprint arXiv:2010.11929}, 2020.

\bibitem{huang2018urban}
B.~Huang, B.~Zhao, and Y.~Song, ``Urban land-use mapping using a deep convolutional neural network with high spatial resolution multispectral remote sensing imagery,'' \emph{Remote Sensing of Environment}, vol. 214, pp. 73--86, 2018.

\bibitem{yang2018hyperspectral}
X.~Yang, Y.~Ye, X.~Li, R.~Y. Lau, X.~Zhang, and X.~Huang, ``Hyperspectral image classification with deep learning models,'' \emph{IEEE Transactions on Geoscience and Remote Sensing}, vol.~56, no.~9, pp. 5408--5423, 2018.

\bibitem{chen2016target}
S.~Chen, H.~Wang, F.~Xu, and Y.-Q. Jin, ``Target classification using the deep convolutional networks for sar images,'' \emph{IEEE transactions on geoscience and remote sensing}, vol.~54, no.~8, pp. 4806--4817, 2016.

\bibitem{gupta2018social}
A.~Gupta, J.~Johnson, L.~Fei-Fei, S.~Savarese, and A.~Alahi, ``Social gan: Socially acceptable trajectories with generative adversarial networks,'' in \emph{Proceedings of the IEEE conference on computer vision and pattern recognition}, 2018, pp. 2255--2264.

\bibitem{velivckovic2017graph}
P.~Veli{\v{c}}kovi{\'c}, G.~Cucurull, A.~Casanova, A.~Romero, P.~Lio, and Y.~Bengio, ``Graph attention networks,'' \emph{arXiv preprint arXiv:1710.10903}, 2017.

\bibitem{wu2019pointconv}
W.~Wu, Z.~Qi, and L.~Fuxin, ``Pointconv: Deep convolutional networks on 3d point clouds,'' in \emph{Proceedings of the IEEE/CVF Conference on computer vision and pattern recognition}, 2019, pp. 9621--9630.

\bibitem{sadeghian2019sophie}
A.~Sadeghian, V.~Kosaraju, A.~Sadeghian, N.~Hirose, H.~Rezatofighi, and S.~Savarese, ``Sophie: An attentive gan for predicting paths compliant to social and physical constraints,'' in \emph{Proceedings of the IEEE/CVF conference on computer vision and pattern recognition}, 2019, pp. 1349--1358.

\bibitem{hughes2020deep}
L.~H. Hughes, D.~Marcos, S.~Lobry, D.~Tuia, and M.~Schmitt, ``A deep learning framework for matching of sar and optical imagery,'' \emph{ISPRS Journal of Photogrammetry and Remote Sensing}, vol. 169, pp. 166--179, 2020.

\bibitem{li2021deep}
X.~Li, Z.~Du, Y.~Huang, and Z.~Tan, ``A deep translation (gan) based change detection network for optical and sar remote sensing images,'' \emph{ISPRS Journal of Photogrammetry and Remote Sensing}, vol. 179, pp. 14--34, 2021.

\bibitem{yang2018hyperspectralandmultispectral}
J.~Yang, Y.-Q. Zhao, and J.~C.-W. Chan, ``Hyperspectral and multispectral image fusion via deep two-branches convolutional neural network,'' \emph{Remote Sensing}, vol.~10, no.~5, p. 800, 2018.

\bibitem{hang2020classification}
R.~Hang, Z.~Li, P.~Ghamisi, D.~Hong, G.~Xia, and Q.~Liu, ``Classification of hyperspectral and lidar data using coupled cnns,'' \emph{IEEE Transactions on Geoscience and Remote Sensing}, vol.~58, no.~7, pp. 4939--4950, 2020.

\bibitem{zhang2023morphological}
M.~Zhang, W.~Li, X.~Zhao, H.~Liu, R.~Tao, and Q.~Du, ``Morphological transformation and spatial-logical aggregation for tree species classification using hyperspectral imagery,'' \emph{IEEE Transactions on Geoscience and Remote Sensing}, vol.~61, pp. 1--12, 2023.

\bibitem{gao2021hyperspectral}
Y.~Gao, W.~Li, M.~Zhang, J.~Wang, W.~Sun, R.~Tao, and Q.~Du, ``Hyperspectral and multispectral classification for coastal wetland using depthwise feature interaction network,'' \emph{IEEE Transactions on Geoscience and Remote Sensing}, vol.~60, pp. 1--15, 2021.

\bibitem{hong2021multimodal}
D.~Hong, J.~Hu, J.~Yao, J.~Chanussot, and X.~X. Zhu, ``Multimodal remote sensing benchmark datasets for land cover classification with a shared and specific feature learning model,'' \emph{ISPRS Journal of Photogrammetry and Remote Sensing}, vol. 178, pp. 68--80, 2021.

\bibitem{zzy2023}
Z.~Zhang, Z.~Ren, C.~Tao, Y.~Zhang, C.~Peng, and H.~Li, ``Grass: Contrastive learning with gradient-guided sampling strategy for remote sensing image semantic segmentation,'' \emph{IEEE Transactions on Geoscience and Remote Sensing}, vol.~61, pp. 1--14, 2023.

\bibitem{pengjian}
J.~Peng, D.~Ye, B.~Tang, Y.~Lei, Y.~Liu, and H.~Li, ``Lifelong learning with cycle memory networks,'' \emph{IEEE Transactions on Neural Networks and Learning Systems}, pp. 1--14, 2023.

\bibitem{yu2022point}
X.~Yu, L.~Tang, Y.~Rao, T.~Huang, J.~Zhou, and J.~Lu, ``Point-bert: Pre-training 3d point cloud transformers with masked point modeling,'' in \emph{Proceedings of the IEEE/CVF Conference on Computer Vision and Pattern Recognition}, 2022, pp. 19\,313--19\,322.

\bibitem{zeng2023matters}
Y.~Zeng, H.~Zhang, J.~Zheng, J.~Xia, G.~Wei, Y.~Wei, Y.~Zhang, and T.~Kong, ``What matters in training a gpt4-style language model with multimodal inputs?'' \emph{arXiv preprint arXiv:2307.02469}, 2023.

\bibitem{padhi2021tabular}
I.~Padhi, Y.~Schiff, I.~Melnyk, M.~Rigotti, Y.~Mroueh, P.~Dognin, J.~Ross, R.~Nair, and E.~Altman, ``Tabular transformers for modeling multivariate time series,'' in \emph{ICASSP 2021-2021 IEEE International Conference on Acoustics, Speech and Signal Processing (ICASSP)}.\hskip 1em plus 0.5em minus 0.4em\relax IEEE, 2021, pp. 3565--3569.

\bibitem{fang2023eva}
Y.~Fang, W.~Wang, B.~Xie, Q.~Sun, L.~Wu, X.~Wang, T.~Huang, X.~Wang, and Y.~Cao, ``Eva: Exploring the limits of masked visual representation learning at scale,'' in \emph{Proceedings of the IEEE/CVF Conference on Computer Vision and Pattern Recognition}, 2023, pp. 19\,358--19\,369.

\bibitem{shi2023trajectory}
L.~Shi, L.~Wang, S.~Zhou, and G.~Hua, ``Trajectory unified transformer for pedestrian trajectory prediction,'' in \emph{Proceedings of the IEEE/CVF International Conference on Computer Vision}, 2023, pp. 9675--9684.

\bibitem{liu2023spatio}
H.~Liu, Z.~Dong, R.~Jiang, J.~Deng, J.~Deng, Q.~Chen, and X.~Song, ``Spatio-temporal adaptive embedding makes vanilla transformer sota for traffic forecasting,'' in \emph{Proceedings of the 32nd ACM International Conference on Information and Knowledge Management}, 2023, pp. 4125--4129.

\bibitem{zhu2022kst}
J.~Zhu, X.~Han, H.~Deng, C.~Tao, L.~Zhao, P.~Wang, L.~Tao, and H.~Li, ``Kst-gcn: A knowledge-driven spatial-temporal graph convolutional network for traffic forecasting,'' \emph{IEEE Transactions on Intelligent Transportation Systems}, vol.~23, no.~9, pp. 15\,055--15\,065, 2022.

\bibitem{devlin2018bert}
J.~Devlin, ``Bert: Pre-training of deep bidirectional transformers for language understanding,'' \emph{arXiv preprint arXiv:1810.04805}, 2018.

\bibitem{touvron2023llama}
H.~Touvron, L.~Martin, K.~Stone, P.~Albert, A.~Almahairi, Y.~Babaei, N.~Bashlykov, S.~Batra, P.~Bhargava, S.~Bhosale \emph{et~al.}, ``Llama 2: Open foundation and fine-tuned chat models,'' \emph{arXiv preprint arXiv:2307.09288}, 2023.

\bibitem{maas2011learning}
A.~Maas, R.~E. Daly, P.~T. Pham, D.~Huang, A.~Y. Ng, and C.~Potts, ``Learning word vectors for sentiment analysis,'' in \emph{Proceedings of the 49th annual meeting of the association for computational linguistics: Human language technologies}, 2011, pp. 142--150.

\bibitem{husain2019codesearchnet}
H.~Husain, H.-H. Wu, T.~Gazit, M.~Allamanis, and M.~Brockschmidt, ``Codesearchnet challenge: Evaluating the state of semantic code search,'' \emph{arXiv preprint arXiv:1909.09436}, 2019.

\bibitem{vinyals2016matching}
O.~Vinyals, C.~Blundell, T.~Lillicrap, D.~Wierstra \emph{et~al.}, ``Matching networks for one shot learning,'' \emph{Advances in neural information processing systems}, vol.~29, 2016.

\bibitem{sung2018learning}
F.~Sung, Y.~Yang, L.~Zhang, T.~Xiang, P.~H. Torr, and T.~M. Hospedales, ``Learning to compare: Relation network for few-shot learning,'' in \emph{Proceedings of the IEEE conference on computer vision and pattern recognition}, 2018, pp. 1199--1208.

\bibitem{snell2017prototypical}
J.~Snell, K.~Swersky, and R.~Zemel, ``Prototypical networks for few-shot learning,'' \emph{Advances in neural information processing systems}, vol.~30, 2017.

\bibitem{li2020dla}
L.~Li, J.~Han, X.~Yao, G.~Cheng, and L.~Guo, ``Dla-matchnet for few-shot remote sensing image scene classification,'' \emph{IEEE Transactions on Geoscience and Remote Sensing}, vol.~59, no.~9, pp. 7844--7853, 2020.

\bibitem{cheng2021spnet}
G.~Cheng, L.~Cai, C.~Lang, X.~Yao, J.~Chen, L.~Guo, and J.~Han, ``Spnet: Siamese-prototype network for few-shot remote sensing image scene classification,'' \emph{IEEE Transactions on Geoscience and Remote Sensing}, vol.~60, pp. 1--11, 2021.

\bibitem{garcia2017few}
V.~Garcia and J.~Bruna, ``Few-shot learning with graph neural networks,'' \emph{arXiv preprint arXiv:1711.04043}, 2017.

\bibitem{zhang2017cautionary}
S.~Zhang, B.~Guo, A.~Dong, J.~He, Z.~Xu, and S.~X. Chen, ``Cautionary tales on air-quality improvement in beijing,'' \emph{Proceedings of the Royal Society A: Mathematical, Physical and Engineering Sciences}, vol. 473, no. 2205, p. 20170457, 2017.

\bibitem{cheng2017remote}
G.~Cheng, J.~Han, and X.~Lu, ``Remote sensing image scene classification: Benchmark and state of the art,'' \emph{Proceedings of the IEEE}, vol. 105, no.~10, pp. 1865--1883, 2017.

\bibitem{helber2019eurosat}
P.~Helber, B.~Bischke, A.~Dengel, and D.~Borth, ``Eurosat: A novel dataset and deep learning benchmark for land use and land cover classification,'' \emph{IEEE Journal of Selected Topics in Applied Earth Observations and Remote Sensing}, vol.~12, no.~7, pp. 2217--2226, 2019.

\bibitem{pellegrini2009you}
S.~Pellegrini, A.~Ess, K.~Schindler, and L.~Van~Gool, ``You'll never walk alone: Modeling social behavior for multi-target tracking,'' in \emph{2009 IEEE 12th international conference on computer vision}.\hskip 1em plus 0.5em minus 0.4em\relax IEEE, 2009, pp. 261--268.

\bibitem{lerner2007crowds}
A.~Lerner, Y.~Chrysanthou, and D.~Lischinski, ``Crowds by example,'' in \emph{Computer graphics forum}, vol.~26, no.~3.\hskip 1em plus 0.5em minus 0.4em\relax Wiley Online Library, 2007, pp. 655--664.

\bibitem{keydel1996mstar}
E.~R. Keydel, S.~W. Lee, and J.~T. Moore, ``Mstar extended operating conditions: A tutorial,'' \emph{Algorithms for Synthetic Aperture Radar Imagery III}, vol. 2757, pp. 228--242, 1996.

\bibitem{wu20153d}
Z.~Wu, S.~Song, A.~Khosla, F.~Yu, L.~Zhang, X.~Tang, and J.~Xiao, ``3d shapenets: A deep representation for volumetric shapes,'' in \emph{Proceedings of the IEEE conference on computer vision and pattern recognition}, 2015, pp. 1912--1920.

\bibitem{yang2010bag}
Y.~Yang and S.~Newsam, ``Bag-of-visual-words and spatial extensions for land-use classification,'' in \emph{Proceedings of the 18th SIGSPATIAL international conference on advances in geographic information systems}, 2010, pp. 270--279.

\bibitem{Dai2011WHURS19}
D.~Dai and W.~Yang, ``Satellite image classification via two-layer sparse coding with biased image representation,'' \emph{IEEE Transactions on Geoscience and Remote Sensing}, vol.~8, no.~1, pp. 173--176, 2011.

\bibitem{Xia2010WHURS19}
G.-S. Xia, W.~Yang, J.~Delon, Y.~Gousseau, H.~Sun, and H.~MaÎtre, ``Structural high-resolution satellite image indexing,'' Vienna, Austria, 2010.

\bibitem{chang2015shapenet}
A.~X. Chang, T.~Funkhouser, L.~Guibas, P.~Hanrahan, Q.~Huang, Z.~Li, S.~Savarese, M.~Savva, S.~Song, H.~Su \emph{et~al.}, ``Shapenet: An information-rich 3d model repository,'' \emph{arXiv preprint arXiv:1512.03012}, 2015.

\bibitem{uy-scanobjectnn-iccv19}
M.~A. Uy, Q.-H. Pham, B.-S. Hua, D.~T. Nguyen, and S.-K. Yeung, ``Revisiting point cloud classification: A new benchmark dataset and classification model on real-world data,'' in \emph{International Conference on Computer Vision (ICCV)}, 2019.

\bibitem{ye2023closer}
C.~Ye, H.~Zhu, B.~Zhang, and T.~Chen, ``A closer look at few-shot 3d point cloud classification,'' \emph{International Journal of Computer Vision}, vol. 131, no.~3, pp. 772--795, 2023.

\bibitem{zhang2019remote}
W.~Zhang, P.~Tang, and L.~Zhao, ``Remote sensing image scene classification using cnn-capsnet,'' \emph{Remote Sensing}, vol.~11, no.~5, p. 494, 2019.

\bibitem{xu2021deep}
K.~Xu, H.~Huang, P.~Deng, and Y.~Li, ``Deep feature aggregation framework driven by graph convolutional network for scene classification in remote sensing,'' \emph{IEEE Transactions on Neural Networks and Learning Systems}, vol.~33, no.~10, pp. 5751--5765, 2021.

\bibitem{cheng2018deep}
G.~Cheng, C.~Yang, X.~Yao, L.~Guo, and J.~Han, ``When deep learning meets metric learning: Remote sensing image scene classification via learning discriminative cnns,'' \emph{IEEE transactions on geoscience and remote sensing}, vol.~56, no.~5, pp. 2811--2821, 2018.

\bibitem{he2019skip}
N.~He, L.~Fang, S.~Li, J.~Plaza, and A.~Plaza, ``Skip-connected covariance network for remote sensing scene classification,'' \emph{IEEE transactions on neural networks and learning systems}, vol.~31, no.~5, pp. 1461--1474, 2019.

\bibitem{manas2021seasonal}
O.~Manas, A.~Lacoste, X.~Gir{\'o}-i Nieto, D.~Vazquez, and P.~Rodriguez, ``Seasonal contrast: Unsupervised pre-training from uncurated remote sensing data,'' in \emph{Proceedings of the IEEE/CVF International Conference on Computer Vision}, 2021, pp. 9414--9423.

\bibitem{wang2020multi}
S.~Wang, Y.~Guan, and L.~Shao, ``Multi-granularity canonical appearance pooling for remote sensing scene classification,'' \emph{IEEE Transactions on Image Processing}, vol.~29, pp. 5396--5407, 2020.

\bibitem{bi2021local}
Q.~Bi, K.~Qin, H.~Zhang, and G.-S. Xia, ``Local semantic enhanced convnet for aerial scene recognition,'' \emph{IEEE Transactions on Image Processing}, vol.~30, pp. 6498--6511, 2021.

\bibitem{zhang2021multiscale}
G.~Zhang, W.~Xu, W.~Zhao, C.~Huang, E.~N. Yk, Y.~Chen, and J.~Su, ``A multiscale attention network for remote sensing scene images classification,'' \emph{IEEE Journal of Selected Topics in Applied Earth Observations and Remote Sensing}, vol.~14, pp. 9530--9545, 2021.

\bibitem{wang2020invariant}
S.~Wang, Y.~Ren, G.~Parr, Y.~Guan, and L.~Shao, ``Invariant deep compressible covariance pooling for aerial scene categorization,'' \emph{IEEE Transactions on Geoscience and Remote Sensing}, vol.~59, no.~8, pp. 6549--6561, 2020.

\bibitem{chen2021remote}
S.-B. Chen, Q.-S. Wei, W.-Z. Wang, J.~Tang, B.~Luo, and Z.-Y. Wang, ``Remote sensing scene classification via multi-branch local attention network,'' \emph{IEEE Transactions on Image Processing}, vol.~31, pp. 99--109, 2021.

\bibitem{li2021gated}
B.~Li, Y.~Guo, J.~Yang, L.~Wang, Y.~Wang, and W.~An, ``Gated recurrent multiattention network for vhr remote sensing image classification,'' \emph{IEEE Transactions on Geoscience and Remote Sensing}, vol.~60, pp. 1--13, 2021.

\bibitem{zhao2023emscnet}
Y.~Zhao, J.~Liu, J.~Yang, and Z.~Wu, ``Emscnet: Efficient multisample contrastive network for remote sensing image scene classification,'' \emph{IEEE Transactions on Geoscience and Remote Sensing}, vol.~61, pp. 1--14, 2023.

\bibitem{wang2022advancing}
D.~Wang, Q.~Zhang, Y.~Xu, J.~Zhang, B.~Du, D.~Tao, and L.~Zhang, ``Advancing plain vision transformer toward remote sensing foundation model,'' \emph{IEEE Transactions on Geoscience and Remote Sensing}, vol.~61, pp. 1--15, 2022.

\bibitem{nampally2023performance}
T.~Nampally, J.~Wu, and S.~Dev, ``Performance comparison of multispectral channels for land use classification,'' in \emph{IGARSS 2023-2023 IEEE International Geoscience and Remote Sensing Symposium}.\hskip 1em plus 0.5em minus 0.4em\relax IEEE, 2023, pp. 6178--6181.

\bibitem{zhang2020semi}
K.~Zhang and H.~Yang, ``Semi-supervised multi-spectral land cover classification with multi-attention and adaptive kernel,'' in \emph{2020 IEEE International Conference on Image Processing (ICIP)}.\hskip 1em plus 0.5em minus 0.4em\relax IEEE, 2020, pp. 1881--1885.

\bibitem{gomez2021msmatch}
P.~G{\'o}mez and G.~Meoni, ``Msmatch: semisupervised multispectral scene classification with few labels,'' \emph{IEEE Journal of Selected Topics in Applied Earth Observations and Remote Sensing}, vol.~14, pp. 11\,643--11\,654, 2021.

\bibitem{goyal2022vision}
P.~Goyal, Q.~Duval, I.~Seessel, M.~Caron, I.~Misra, L.~Sagun, A.~Joulin, and P.~Bojanowski, ``Vision models are more robust and fair when pretrained on uncurated images without supervision,'' \emph{arXiv preprint arXiv:2202.08360}, 2022.

\bibitem{wanyan2023dino}
X.~Wanyan, S.~Seneviratne, S.~Shen, and M.~Kirley, ``Dino-mc: Self-supervised contrastive learning for remote sensing imagery with multi-sized local crops,'' \emph{arXiv preprint arXiv:2303.06670}, 2023.

\bibitem{uchaev2023small}
D.~Uchaev and D.~Uchaev, ``Small sample hyperspectral image classification based on the random patches network and recursive filtering,'' \emph{Sensors}, vol.~23, no.~5, p. 2499, 2023.

\bibitem{xie2020evolving}
H.~Xie, L.~Zhang, and C.~P. Lim, ``Evolving cnn-lstm models for time series prediction using enhanced grey wolf optimizer,'' \emph{IEEE access}, vol.~8, pp. 161\,519--161\,541, 2020.

\bibitem{cheng2022stacked}
X.~Cheng, W.~Zhang, A.~Wenzel, and J.~Chen, ``Stacked resnet-lstm and coral model for multi-site air quality prediction,'' \emph{Neural Computing and Applications}, vol.~34, no.~16, pp. 13\,849--13\,866, 2022.

\bibitem{li2022prediction}
D.~Li, J.~Liu, and Y.~Zhao, ``Prediction of multi-site pm2. 5 concentrations in beijing using cnn-bi lstm with cbam,'' \emph{Atmosphere}, vol.~13, no.~10, p. 1719, 2022.

\bibitem{guo2020graphcodebert}
D.~Guo, S.~Ren, S.~Lu, Z.~Feng, D.~Tang, S.~Liu, L.~Zhou, N.~Duan, A.~Svyatkovskiy, S.~Fu \emph{et~al.}, ``Graphcodebert: Pre-training code representations with data flow,'' \emph{arXiv preprint arXiv:2009.08366}, 2020.

\bibitem{klokov2017escape}
R.~Klokov and V.~Lempitsky, ``Escape from cells: Deep kd-networks for the recognition of 3d point cloud models,'' in \emph{Proceedings of the IEEE international conference on computer vision}, 2017, pp. 863--872.

\bibitem{lei2020spherical}
H.~Lei, N.~Akhtar, and A.~Mian, ``Spherical kernel for efficient graph convolution on 3d point clouds,'' \emph{IEEE transactions on pattern analysis and machine intelligence}, vol.~43, no.~10, pp. 3664--3680, 2020.

\bibitem{qi2017pointnet++}
C.~R. Qi, L.~Yi, H.~Su, and L.~J. Guibas, ``Pointnet++: Deep hierarchical feature learning on point sets in a metric space,'' \emph{Advances in neural information processing systems}, vol.~30, 2017.

\bibitem{li2018so}
J.~Li, B.~M. Chen, and G.~H. Lee, ``So-net: Self-organizing network for point cloud analysis,'' in \emph{Proceedings of the IEEE conference on computer vision and pattern recognition}, 2018, pp. 9397--9406.

\bibitem{li2021pointvgg}
R.~Li, Y.~Zhang, D.~Niu, G.~Yang, N.~Zafar, C.~Zhang, and X.~Zhao, ``Pointvgg: Graph convolutional network with progressive aggregating features on point clouds,'' \emph{Neurocomputing}, vol. 429, pp. 187--198, 2021.

\bibitem{chen2023pointgpt}
G.~Chen, M.~Wang, Y.~Yang, K.~Yu, L.~Yuan, and Y.~Yue, ``Pointgpt: Auto-regressively generative pre-training from point clouds,'' \emph{arXiv preprint arXiv:2305.11487}, 2023.

\bibitem{srinivas2014sar}
U.~Srinivas, V.~Monga, and R.~G. Raj, ``Sar automatic target recognition using discriminative graphical models,'' \emph{IEEE transactions on aerospace and electronic systems}, vol.~50, no.~1, pp. 591--606, 2014.

\bibitem{dong2014sparse}
G.~Dong, N.~Wang, and G.~Kuang, ``Sparse representation of monogenic signal: With application to target recognition in sar images,'' \emph{IEEE signal processing letters}, vol.~21, no.~8, pp. 952--956, 2014.

\bibitem{dong2015classification}
G.~Dong and G.~Kuang, ``Classification on the monogenic scale space: Application to target recognition in sar image,'' \emph{IEEE Transactions on Image Processing}, vol.~24, no.~8, pp. 2527--2539, 2015.

\bibitem{o2001sar}
J.~A. O'Sullivan, M.~D. DeVore, V.~Kedia, and M.~I. Miller, ``Sar atr performance using a conditionally gaussian model,'' \emph{IEEE Transactions on Aerospace and Electronic Systems}, vol.~37, no.~1, pp. 91--108, 2001.

\bibitem{park2014modified}
J.-I. Park and K.-T. Kim, ``Modified polar mapping classifier for sar automatic target recognition,'' \emph{IEEE Transactions on Aerospace and Electronic Systems}, vol.~50, no.~2, pp. 1092--1107, 2014.

\bibitem{liu2019roberta}
Y.~Liu, M.~Ott, N.~Goyal, J.~Du, M.~Joshi, D.~Chen, O.~Levy, M.~Lewis, L.~Zettlemoyer, and V.~Stoyanov, ``Roberta: A robustly optimized bert pretraining approach,'' \emph{arXiv preprint arXiv:1907.11692}, 2019.

\bibitem{howard2018universal}
J.~Howard and S.~Ruder, ``Universal language model fine-tuning for text classification,'' \emph{arXiv preprint arXiv:1801.06146}, 2018.

\bibitem{xie2020unsupervised}
Q.~Xie, Z.~Dai, E.~Hovy, T.~Luong, and Q.~Le, ``Unsupervised data augmentation for consistency training,'' \emph{Advances in neural information processing systems}, vol.~33, pp. 6256--6268, 2020.

\bibitem{sachan2019revisiting}
D.~S. Sachan, M.~Zaheer, and R.~Salakhutdinov, ``Revisiting lstm networks for semi-supervised text classification via mixed objective function,'' in \emph{Proceedings of the AAAI Conference on Artificial Intelligence}, vol.~33, no.~01, 2019, pp. 6940--6948.

\bibitem{miyato2016adversarial}
T.~Miyato, A.~M. Dai, and I.~Goodfellow, ``Adversarial training methods for semi-supervised text classification,'' \emph{arXiv preprint arXiv:1605.07725}, 2016.

\bibitem{beltagy2020longformer}
I.~Beltagy, M.~E. Peters, and A.~Cohan, ``Longformer: The long-document transformer,'' \emph{arXiv preprint arXiv:2004.05150}, 2020.

\bibitem{yang2019xlnet}
Z.~Yang, Z.~Dai, Y.~Yang, J.~Carbonell, R.~R. Salakhutdinov, and Q.~V. Le, ``Xlnet: Generalized autoregressive pretraining for language understanding,'' \emph{Advances in neural information processing systems}, vol.~32, 2019.

\bibitem{ding2020ernie}
S.~Ding, J.~Shang, S.~Wang, Y.~Sun, H.~Tian, H.~Wu, and H.~Wang, ``Ernie-doc: A retrospective long-document modeling transformer,'' \emph{arXiv preprint arXiv:2012.15688}, 2020.

\bibitem{Cui2024AEPT}
J.~Cui, W.~Guo, H.~Huang, X.~Lv, H.~Cao, and H.~Li, ``Adversarial examples for vehicle detection with projection transformation,'' \emph{IEEE Transactions on Geoscience and Remote Sensing}, vol.~62, p. 5632418, 2024.

\bibitem{soomro2012ucf101}
K.~Soomro, A.~R. Zamir, and M.~Shah, ``Ucf101: A dataset of 101 human actions classes from videos in the wild,'' \emph{arXiv preprint arXiv:1212.0402}, 2012.

\bibitem{liu2023deep}
J.~Liu, J.~Gao, S.~Ji, C.~Zeng, S.~Zhang, and J.~Gong, ``Deep learning based multi-view stereo matching and 3d scene reconstruction from oblique aerial images,'' \emph{ISPRS Journal of Photogrammetry and Remote Sensing}, vol. 204, pp. 42--60, 2023.

\end{thebibliography}

\end{document}